\RequirePackage{fix-cm}
\documentclass{svjour3}                     
\smartqed  
\usepackage{graphicx}
\usepackage{multirow}
\usepackage[misc]{ifsym}
\usepackage{booktabs}
\usepackage{subfigure}
\usepackage{natbib}
\usepackage{color}
\usepackage{amsmath,amsfonts}
\usepackage[colorlinks,linkcolor=blue,anchorcolor=blue,citecolor=blue,urlcolor=blue]{hyperref}

\usepackage{geometry}
\begin{document}

\title{A deep learning method based on patchwise training for reconstructing temperature field
}


\author{Xingwen Peng \textsuperscript{1}   \and
		Xingchen Li \textsuperscript{2,*} \and
        Zhiqiang Gong \textsuperscript{2} \and      
	    Xiaoyu Zhao \textsuperscript{2} \and
        Wen Yao \textsuperscript{2,*} \and
}


\institute{
	\Letter{~Wen Yao} \\
	\email{wendy0782@126.com} \\
	Tel.: +86-18518169621 \\
	Fax: +86-010-870400 \\
	\at
	\textsuperscript{1}College of Aerospace Science and Engineering, National University of Defense Technology, Changsha, 410073, China\\	
	\textsuperscript{2}Defense Innovation Institute, Chinese Academy of Military Science, Beijing 100071, China}

\date{Received: date / Accepted: date}

\maketitle

\begin{abstract}

Physical field reconstruction is highly desirable for the measurement and control of engineering systems. The reconstruction of the temperature field from limited observation plays a crucial role in thermal management for electronic equipment. Deep learning has been employed in physical field reconstruction, whereas the accurate estimation for the regions with large gradients is still diffcult. To solve the problem, this work proposes a novel deep learning method based on patchwise training to reconstruct the temperature field of electronic equipment accurately from limited observation. Firstly, the temperature field reconstruction (TFR) problem of the electronic equipment is modeled mathematically and transformed as an image-to-image regression task. Then a patchwise training and inference framework consisting of an adaptive UNet and a shallow multilayer perceptron (MLP) is developed to establish the mapping from the observation to the temperature field. The adaptive UNet is utilized to reconstruct the whole temperature field while the MLP is designed to predict the patches with large temperature gradients. Experiments employing finite element simulation data are conducted to demonstrate the accuracy of the proposed method. Furthermore, the generalization is evaluated by investigating cases under different heat source layouts, different power intensities, and different observation point locations. The maximum absolute errors of the reconstructed temperature field are less than 1K under the patchwise training approach.
\end{abstract}

\section{Introduction}
\label{intro}
Well thermal management (\cite{Thermal}) is arguably the most crucial goal in electronic equipment design, which requires not only a deep understanding of heat transfer but also good design strategies. The heat is mainly generated by integrated electronic components with small size and high-power density (\cite{HSLO}). Generally, the reliability and service life of electronic component are significantly affected by temperature (\cite{55}). In addition, thermal stress and thermal strain due to the non-uniform temperature distribution may cause solder fatigue and electronic components failure (\cite{Stress, Deformation}). Obtaining temperature distribution information is highly desirable for the thermal management of electronic equipment. The typical temperature measurement techniques generally use transducers (\cite{Transducer}) (e.g. thermocouples and thermistors) to obtain local information, which cannot describe the state of measurement area comprehensively. Generally, more measurement points can provide more spatial distribution information. However, only limited measurement points can be deployed due to restrictions of the engineering practice in a real-world system. Reconstructing temperature field from limited temperature observation, as an inverse heat transfer problem (\cite{Inverse}), is an interesting and challenging task.

Available approaches to reconstruct the temperature field fall into two main categories. The first one directly uses numerical methods such as interpolation (\cite{Spline, IDW}) and iteration (\cite{Iteration}), which reconstruct the target temperature of points or domains within the range of a discrete set of known temperature observation points. Generally speaking, these methods are easy to be implemented but need a substantial number of known temperature observations to obtain high precision and are usually incredibly time-consuming. The other one takes advantage of the surrogate model-based methods to cope with those problems. The methods concerned about TFR is teeming with polynomial regression (\cite{PR}), kriging (\cite{kriging}), radial basis function(RBF) (\cite{RBF}), support vector regression(SVR) (\cite{SVR}), gappy proper orthogonal decomposition (GPOD) (\cite{GPOD}), Kalman filtering (\cite{Kalman}), Extreme Learning Machine(ELM) (\cite{ELM}), neural network(NN) (\cite{NN, Shallow}). However, these traditional surrogate model-based methods can hardly solve the high dimensional nonlinear problems well subject to a limited number of parameters, which means limited representation ability. Thereby choosing a surrogate model with high dimensional and nonlinear correlation representation capacity is vitally important for the TFR task.

In recent years, deep learning technology has achieved great success in various fields such as image classification (\cite{Classification}), image reconstruction (\cite{Reconstruction}), object detection (\cite{Detection}), image super-resolution (\cite{ImageSR}) and semantic segmentation (\cite{Segmentation}) for its powerful feature extraction ability and non-linear mapping capability. For physical problems,  deep learning can be applied to construct surrogate models to learn the physical correlation where the actual physical model is complicated or even unknown (\cite{Transfer}). As powerful universal approximators, deep neural networks (DNNs) have been extended to various physical field reconstructions. \cite{Nano} presented a supervised learning method for the physical field reconstruction in a nanofluid heat transfer problem. \cite{TFRHSS} developed a novel physics-informed deep reversible regression model for TFR of heat-source systems. \cite{Generalization} applied convolutional neural network (CNN) to a number of flowfield reconstruction problems in fluid dynamics. \cite{BNN} proposed a physics-constrained Bayesian deep learning approach to reconstruct flow fields from sparse, noisy velocity data. \cite{RNN} reconstructed the temporal behavior of turbulent flows using a recurrent neural network (RNN). \cite{GAN} applied super-resolution generative adversarial networks (GANs) as a methodology for the reconstruction of turbulent-flow quantities from coarse wall measurements. \cite{PINN} reconstructed velocity and pressure fields from only visualizable concentration data based on a physics-informed Neural Network (PINN) framework. These studies suggest that deep learning methods have great potentials for broader areas of physical field reconstruction. However, there still exist some difficulties faced in applying deep learning methods in the TFR task. For instance, it is hard to guarantee reconstruction accuracy based on a few observations. Besides, large gradient is a common characteristic due to the underlying mechanisms and complex boundary conditions for a lot of physical phenomena, including heat transfer and fluid flow. Unfortunately, deep learning-based methods always provide inaccurate estimations for regions with large gradients. 

To address these problems, this work proposes a deep learning method based on patchwise training to reconstruct the temperature field from limited observation. First, training data is generated via the finite element method (FEM) (\cite{FEM2}). Second, the observation and the temperature field of the domain are regarded as input and output images, respectively. Hence, the TFR can be treated as an image-to-image regression task. Then a patchwise training and inference framework consisting of an adaptive UNet (\cite{Unet}) and a shallow MLP (\cite{MLP}) is developed to learn the underlying laws and establish a mapping from the observation to the temperature field. This overcomes the high-dimensional input and output challenge faced by traditional surrogate model-based methods for the TFR task. In particular, the shallow MLP is designed to improve the reconstruction accuracy of regions with large gradients. Besides, extensive experiments are successfully conducted to demonstrate the accuracy and generalization of the proposed method by investigating cases under different heat source layouts, different power intensities, and different locations of observation points.

On the whole, the main contributions of this work are as follows.
\begin{enumerate}
	\item [(1)] The deep learning method is introduced to cope with the temperature field reconstructon of electronic equipment and, latterly, is provided with mathematical formulation and numerical model.
	\item [(2)] A patchwise training and inference framework consisting of an adaptive UNet and a shallow MLP is developed to improve the reconstruction accuracy of regions with large gradients.
    \item [(3)] Experiments employing simulation data are conducted to demonstrate the accuracy and generalization of the proposed method by investigating cases under different heat source layouts, different power intensities, and different locations of observation points.
\end{enumerate}
This article is organized as follows. In Section \ref{sec:2}, the mathematical model of the TFR problem is established. Then the framework of the proposed method, data preparation, observation points representation, the architecture of deep neural networks, loss function, training and inference are introduced in Section \ref{sec:3}. The experimental setups are described in Section \ref{sec:4}. Next, the experimental studies are presented to validate the effectiveness and generalization of the proposed method in Section \ref{sec:5}. Finally, we conclude this work with some discussions in Section \ref{sec:6}.

\section{Mathematical modeling for TFR problem}
\label{sec:2}
\begin{figure}[htbp]
	\centering
	\includegraphics[width=0.38\linewidth]{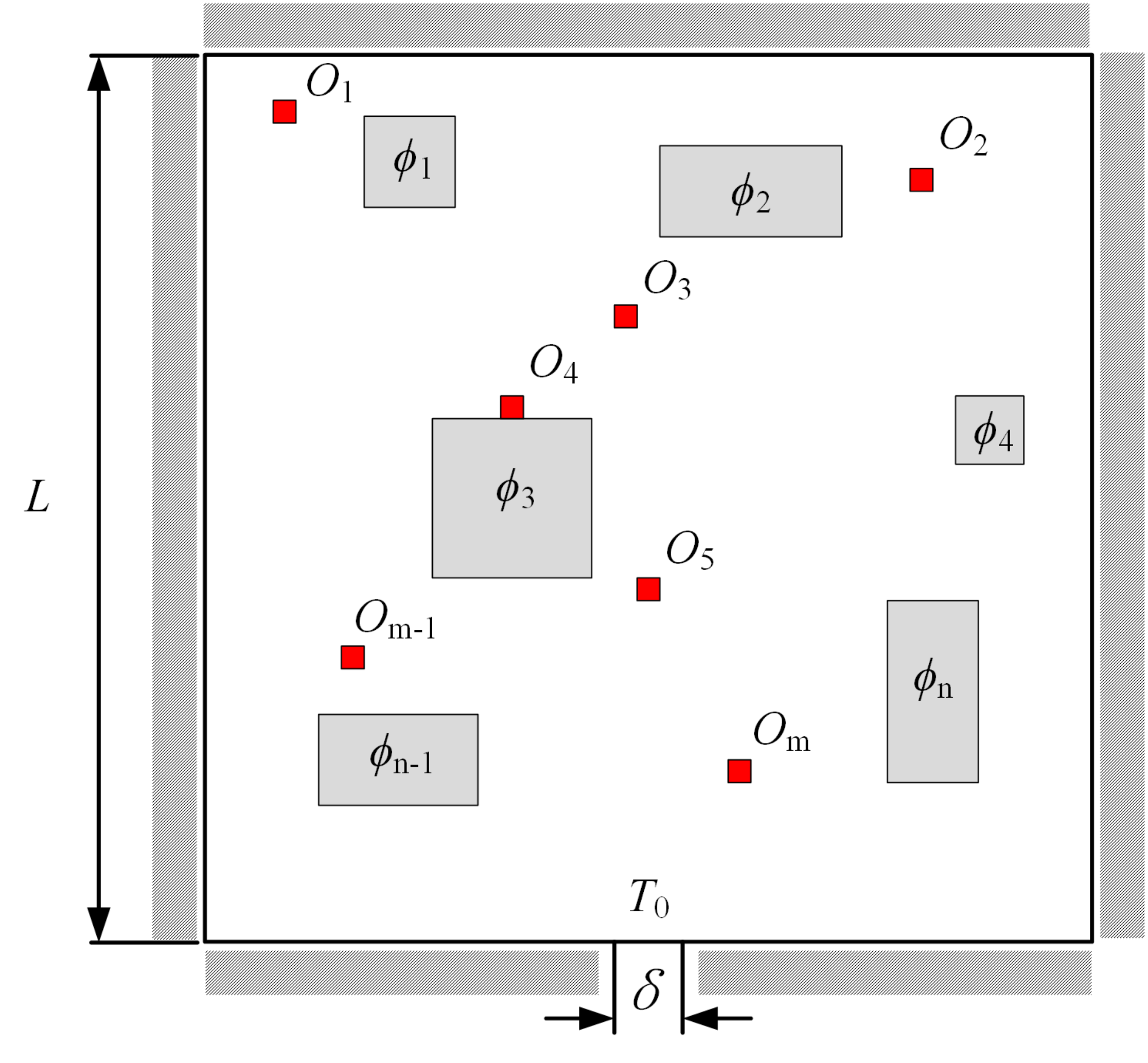}
	\caption{The illustration of the VP problem with n heat source and m temperature observation points on a domain Ω.}
	\label{fig:vp_problem}
\end{figure}

In the present paper, the TFR problem of electronic equipment for which each component can be seen as a rectangular heat source is studied. The TFR aims to reconstruct the whole temperature field of electronic equipment from limited temperature observation. An electronic equipment system is modeled as a two-dimensional rectangular domain with many rectangular heat sources fixed on it. Without loss of generality, the volume-to-point(VP) problem(\cite{VP}) as Fig. \ref{fig:vp_problem} shows is taken as an example to validate the proposed method. There are $n$ heat sources and $m$ temperature observation points are placed on the layout domain.  $\phi_i(i=1,2,\cdots,n)$ is the power intensity of the $i$th heat source, and $O_1,O_2,\cdots,O_m$ denote the temperature observation points. In addition, all the boundaries are adiabatic except a small patch which is called heat sink ($T_0$) and represented as $\delta$ in the middle of the bottom boundary.
Generally, the steady-state temperature field for two-dimensional VP problems satisfies Poisson’s equation, which can be formulated as
\begin{equation}
\frac{\partial }{{\partial x}}\left( {\lambda\frac{{\partial T}}{{\partial x}}} \right) + \frac{\partial }{{\partial y}}\left( {\lambda\frac{{\partial T}}{{\partial y}}} \right) + \sum\limits_{i=1}^n \phi_i(x,y) = 0
\end{equation}
where $\lambda$ represents the thermal conductivity of the domain. 
For simplicity, the power intensity of each heat source is uniformly distributed and set to be a constant value. Thus it can be described as
\begin{equation}
	\phi_i (x,y) = \left\{ {\begin{array}{*{20}{c}}
			{{\phi _i},}&{(x,y) \in \Gamma }\\
			{0,}&{(x,y) \notin \Gamma }
	\end{array}} \right.
\end{equation}
where $\Gamma$ denotes the area covered by the heat source.

It should also be constrained by the boundary conditions, which can be written as
\begin{equation}
{\rm{Boundary}}:T = {T_0}{\rm{~or~}}k\frac{{\partial T}}{{\partial n}} = 0{\rm{~or~}}k\frac{{\partial T}}{{\partial n}} = h(T - {T_0})
\end{equation}

where $T_0$ is a constant temperature value, $n$ denotes the(typically exterior) normal to the boundary, and $h$ represents the convective heat transfer coefficient. The three boundaries are known as Dirichlet boundary conditions (Dirichlet BCs) where $T_0$ is the isothermal boundary temperature value, Neumann boundary conditions (Neumann BCs) where zero heat flux is exchanged, and Robin boundary conditions (Robin BCs)  where $T_0$ denotes the surrounded fluid temperature value.
Overall, the mathematical  formulation of the TFR problem can be formulated as

\begin{equation}
\left\{\begin{array}{lll}
\textup { find } & \hat{T} \\
\textup { min } & \left|\hat{T}-T\right| \\
\textup { s.t. } & \frac{\partial}{\partial x}\left(\lambda \frac{\partial T}{\partial x}\right)+\frac{\partial}{\partial y}\left(\lambda \frac{\partial T}{\partial y}\right)+\sum_{i=1}^n \phi_{i}(x, y)=0 \\
& T=T_{0} \textup { or } \lambda \frac{\partial T}{\partial \mathbf{n}}=0 \textup { or } \lambda \frac{\partial T}{\partial \mathbf{n}}=h\left(T-T_{0}\right) \\
& T_{i,j, k}=f(i)(i=1,2,\cdots,m) \\
\end{array}\right.
\end{equation}

where $\hat{T}$ stands for the reconstructed temperature by the deep learning surrogate model and $T$ stands for the real temperature field. $T_{i, j, k}$ and $(j, k)$ describe the value and position of the $O_i$ temperature observation point. The objective of the TFR problem is to reconstruct the temperature field from the $m$ temperature observation points.

\section{Using deep learning to reconstruct temperature field}
\label{sec:3}

\begin{figure*}[htbp]
	\centering
	\includegraphics[width=0.96\linewidth]{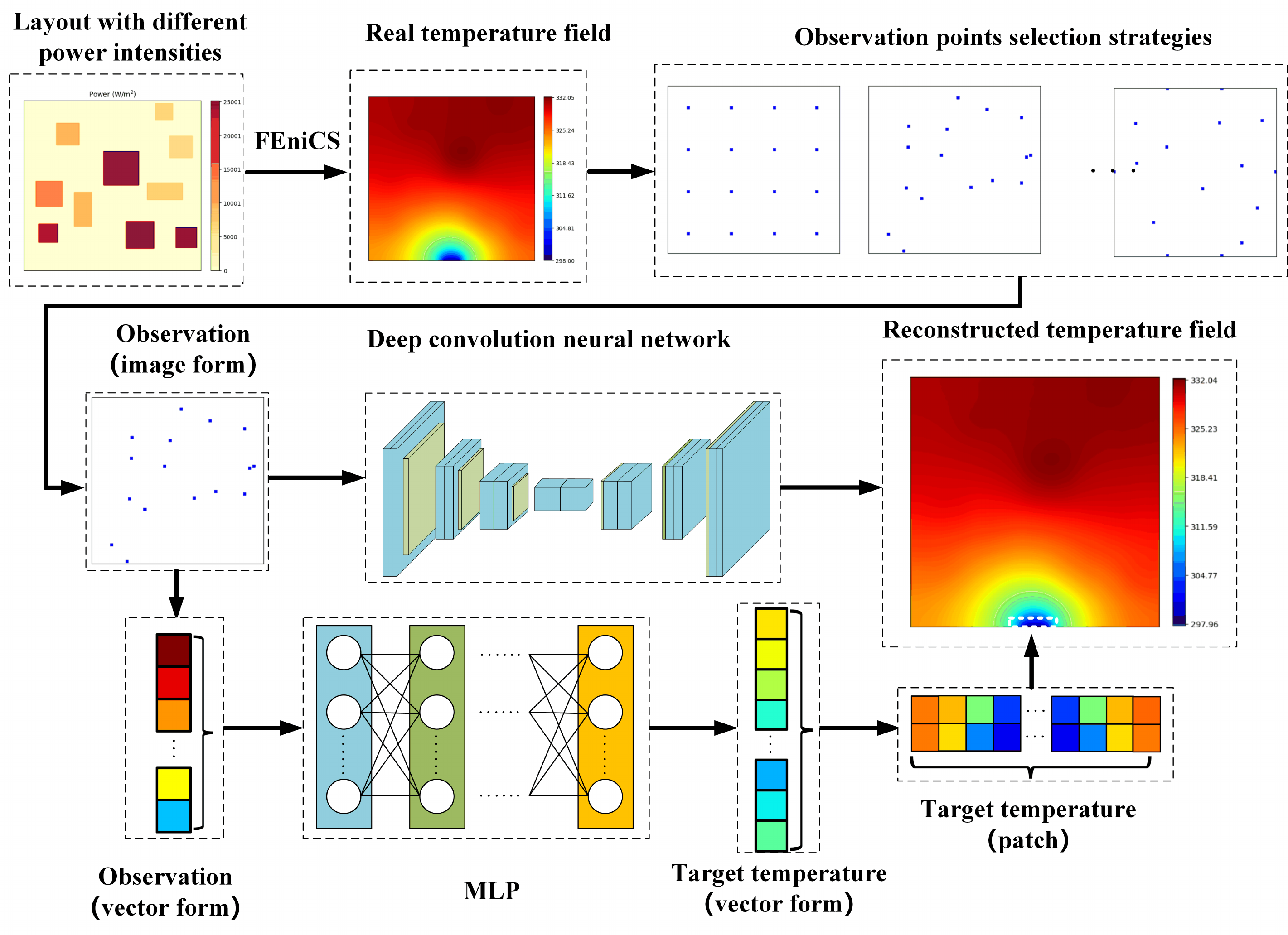}
	\caption{The framework of proposed surrogate model for the TFR problem.}
	\label{fig:data_driven}
\end{figure*}

In what follows, a deep learning method based on patchwise training is proposed to reconstruct the temperature field from limited observation. Fig. \ref{fig:data_driven} illustrates the framework of the proposed method. Firstly, the real temperature field of a heat source layout with different power intensities is generated via a data generator provided by \cite{TFRD}. Secondly, observation points are determined through different selection strategies from the real temperature field and converted to image form and vector form, respectively. Then a mapping between the limited observation (image form) and the temperature field is established through an adaptive UNet. Meanwhile, the observation (vector form) is used as input of a shallow MLP to train and predict the temperature of the target region near the heatsink. The temperature of the target region can be denoted as a patch to replace the temperature estimated by the adaptive UNet (see subsection \ref{sec:5.1.2} for more details). At last, a well-trained surrogate model including a deep CNN and a shallow MLP can reconstruct the overall temperature field given limited observation.

\subsection{Data Preparation}
\label{sec:3.1}

This work denotes the TFR as an image-to-image regression problem using the deep learning surrogate model. To simplify the problem, the domain is set as a square one with the size $L=0.1m$. A certain number of rectangle heat sources with different sizes (e.g., $l_i\times w_i$) are fixed in the domain and the power intensity $\phi_i(i=1,2,...,n)$ of each heat source ranges from $0$ to $30000W/m^2$. The heat generated by the heat source is taken away through the heat sink, for which the length is set to $\delta=0.01m$ and the temperature value is set as constant at $T_0$=298K. 
\begin{figure*}[h]
	\centering
	\subfigure[Heat source layout]{
		\centering
		\includegraphics[width=0.30\linewidth]{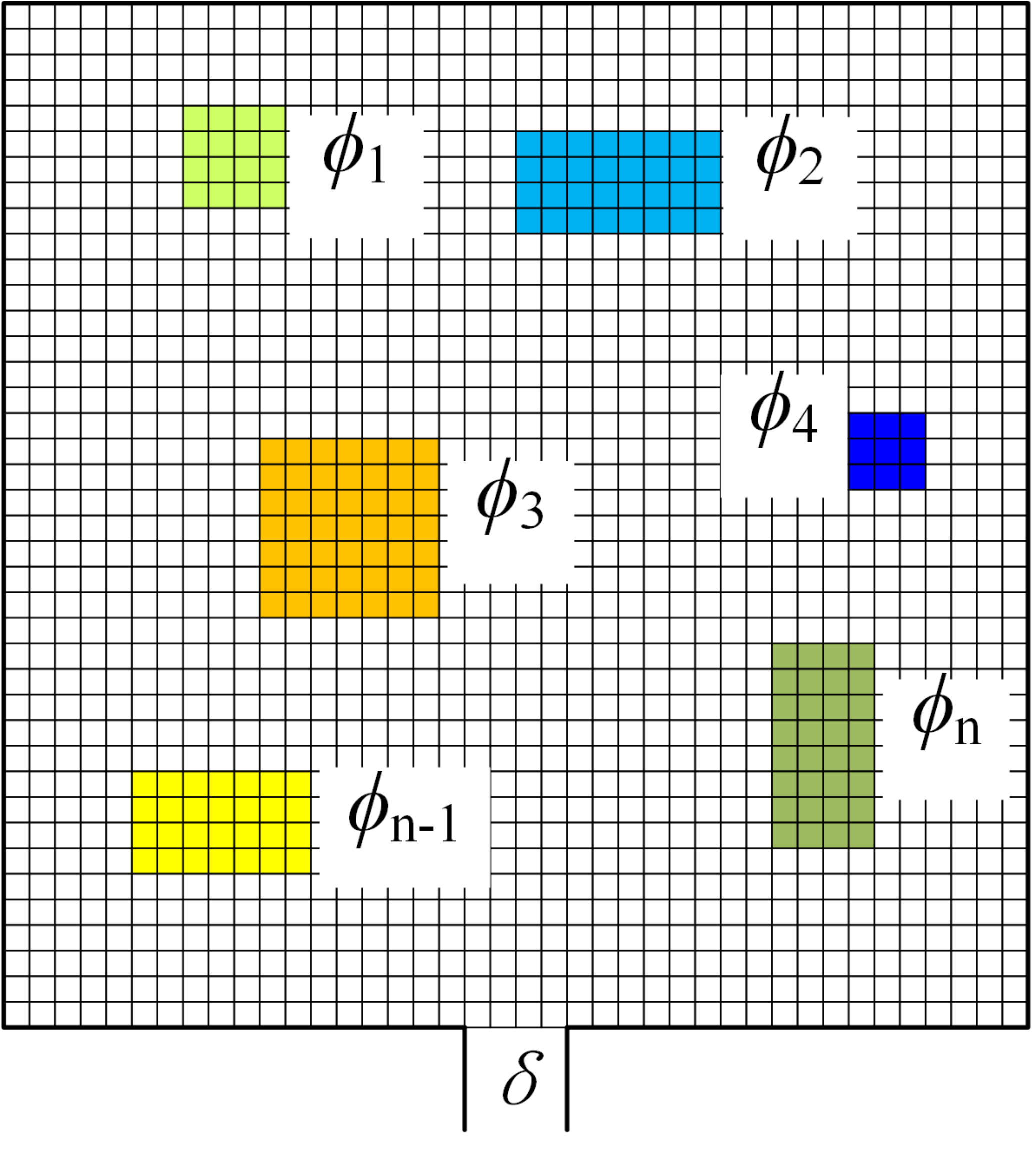}
		\label{fig:Numerical_model_a}
	}
	\subfigure[Temperature field]{
		\centering
		\includegraphics[width=0.30\linewidth]{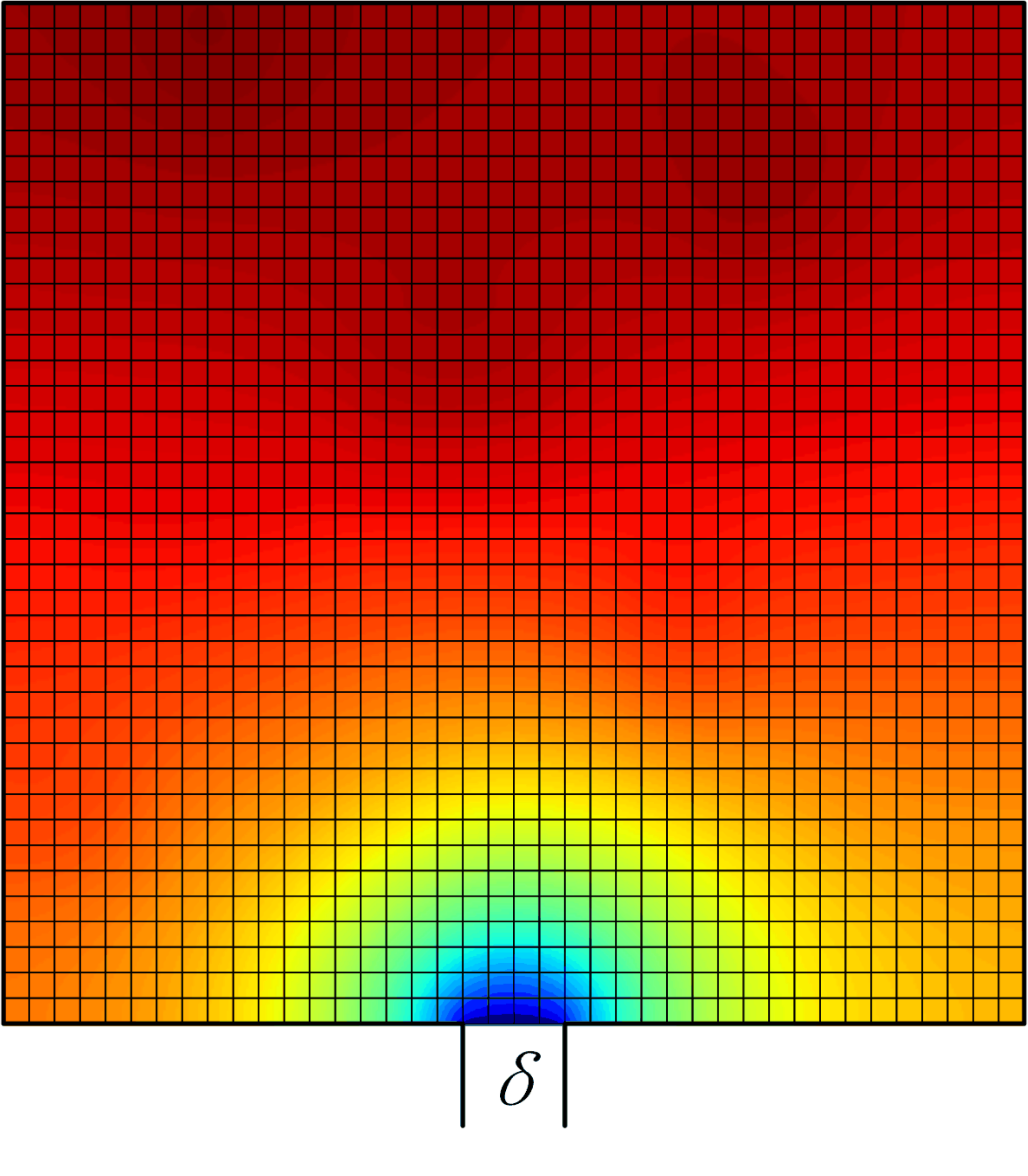}
		\label{fig:Numerical_model_b}
	}
	\subfigure[Observation points]{
		\centering
		\includegraphics[width=0.30\linewidth]{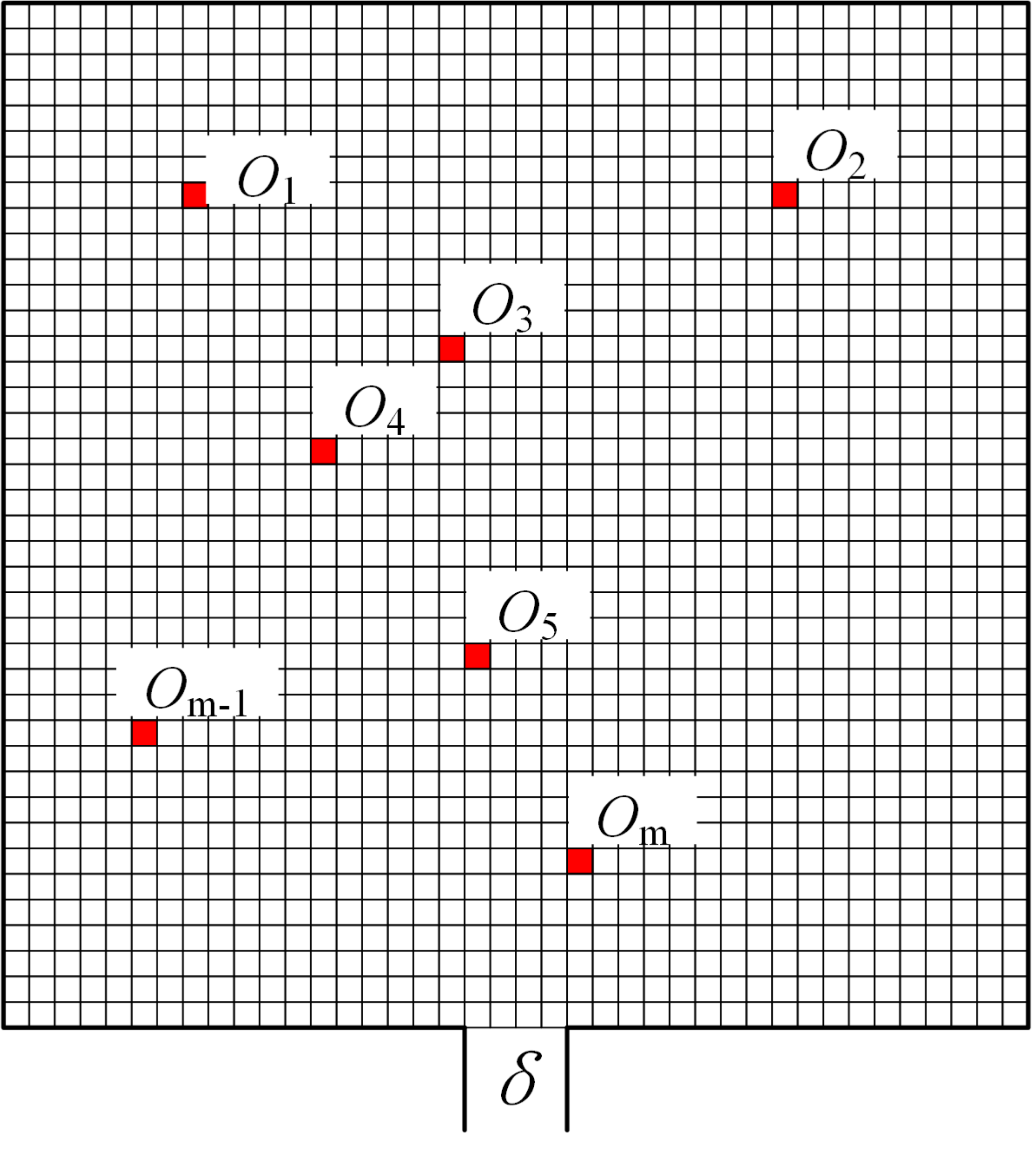}
		\label{fig:Numerical_model_c}
	}
	\caption{Numerical modelling of TFR problem}
	\label{fig:Numerical_model}
\end{figure*}

For the convenience of calculations, the domain is meshed as a $200\times200$ grid system just as Fig. \ref{fig:Numerical_model_a} shows. As a matter of fact, the area with heat sources laid on can be easily represented by a $l_i\times w_i$ grid matrix, where the value of one grid cell is $\phi_i$. On the other hand, the value of one grid cell is 0 for the area without heat sources laid on. Then, FEniCS, a popular open-source computing platform for solving partial differential equations (PDEs) based on FEM, is used as the core solver of a data generator provided by \cite{TFRD} to generate numerical results which are regarded as the real temperature field. In detail, the corresponding temperature field varies alone with the power intensity of each heat source changes. As a result, the $200\times200$ temperature field matrix $T$ as Fig. \ref{fig:Numerical_model_b} shows can be obtained and used as a label in the supervised learning process.

Considering the real-world situation that temperature transducers are arranged fixedly, the locations of temperature observation points are selected through selection strategies (see subsection \ref{sec:5.2.3} for more details) at the beginning and remain unchanged during the training and inference procedure. Then a two-dimensional matrix $T_{ob}$ of $200\times200$ with temperature observation which are the values of fixed grid cells can be acquired. The remainder grid cells of $T_{ob}$ will be filled with $0$ as Fig. \ref{fig:Numerical_model_c} shows. At last, the matrix $T_{ob}$ is fed in the deep neural networks to reconstruct the overall temperature field. Consequently, the TFR problem can be transformed into an image-to-image regression problem for both the input and output are $200\times200$ matrices. It is a challenging problem with high-dimensional input and output for traditional surrogate modeling methods. 

\subsection{Architecture of deep neural networks}
\label{sec:3.2}

\begin{figure*}[htbp]
	\centering
	\includegraphics[width=0.90\linewidth]{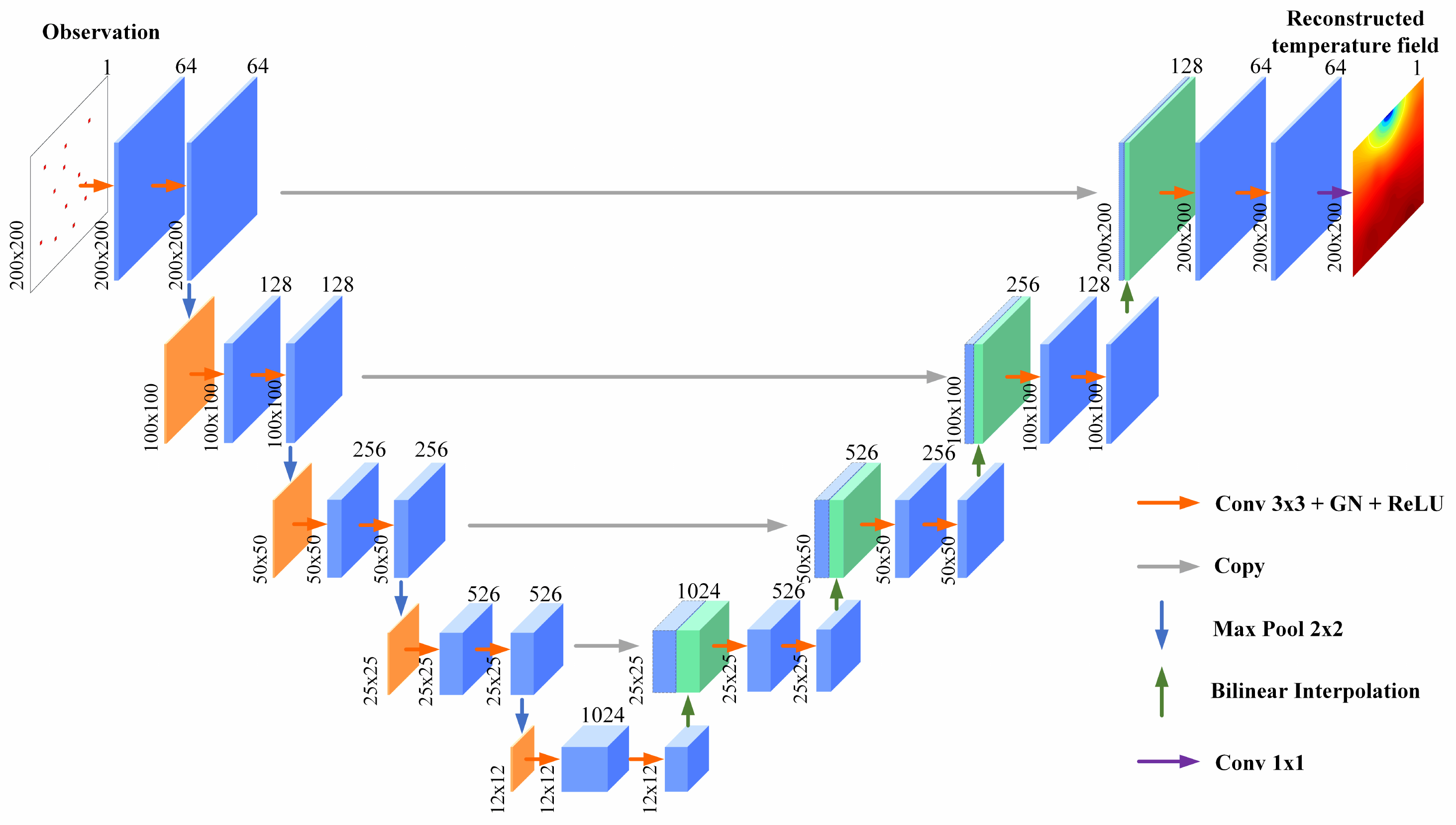}
	\caption{The pipeline of adaptive Unet for TFR.}
	\label{fig:adaptiveUnet}
\end{figure*}

There is large consent that CNN (\cite{CNN}) has made a great success for image-to-image regression tasks. Considering the characteristics of the TFR problem, an adaptive UNet is developed as the main part of the deep learning surrogate model. Unet is an excellent deep CNN framework built upon fully convolutional network (FCN) (\cite{FCN}) and is initially used for biomedical image segmentation. The original UNet architecture consists of a contracting path to capture context and a symmetric expanding path that enables precise localization, which essentially is an encoder-encoder model. Furthermore, UNet takes advantage of the skip connection, which combines high-resolution features from the contracting path with the upsampled output, thus getting a more precise output. Unlike FCN, Unet has plenty of feature channels in the expanding path, which allow the network to propagate context information to higher resolution layers. To acquire better performance, an adaptive Unet is developed as shown in Fig. \ref {fig:adaptiveUnet}, different from the original UNet architecture, the up-convolution operation in the expansive path is replaced by bilinear interpolation to make the reconstructed temperature field smoother. And Group normalization (GN) is adopted after the $3\times3$ convolution operation to obtain a stable accuracy since its computation is independent of batch sizes. 

\begin{figure*}[htbp]
	\centering
	\includegraphics[width=1\linewidth]{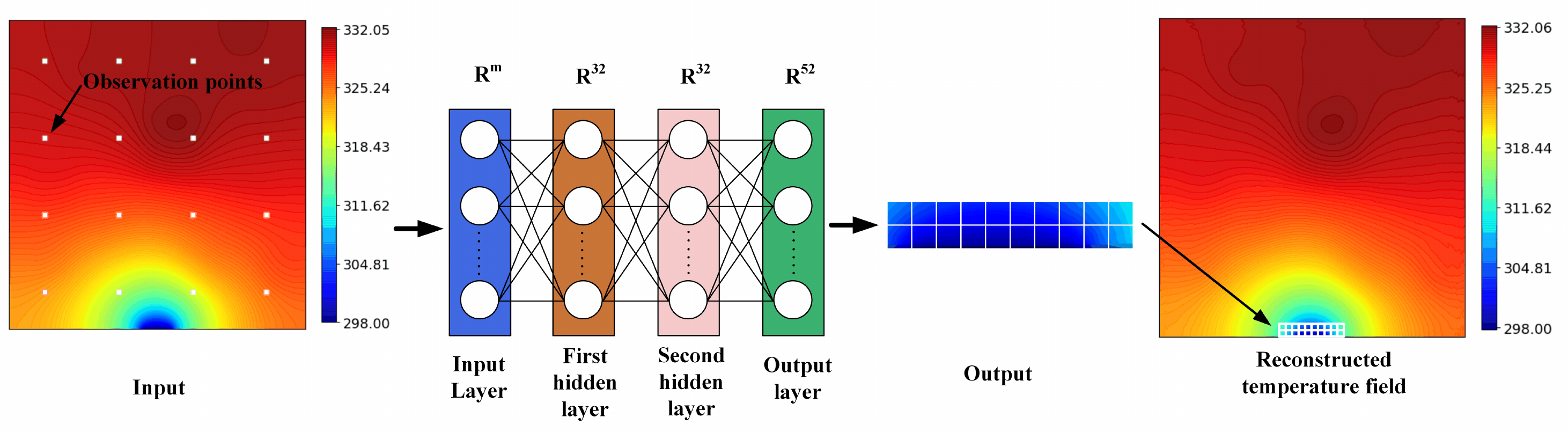}
	\caption{The pipline of MLP for TFR.}
	\label{fig:MLP}
\end{figure*}

Since the temperature gradient is large near the heatsink, the reconstruction performance of pure adaptive UNet is not good in the small area. To alleviate the problem, a viable solution is to take a patchwise training approach, i.e., employing a separate neural network to learn and predict the temperature distribution of the special domain. Considering the computation precision and the efficiency, a shallow MLP consisting of two hidden layers is designed to learn the input-to-output mapping between the observation (vector form) and the temperature of the special domain (patch) as shown in Fig. \ref {fig:MLP}. It should be pointed that this shallow neural network is independent of the adaptive UNet as an assistant method to improve performance. 

\subsection{Loss function and reconstruction}
\label{sec:3.3}

The TFR is essentially a regression task in high-dimensional space. Thus the mean absolute error between the reconstructed temperature field $\hat T$ and the ground-truth temperature field $T$ is chosen as the global field loss function,  which is expressed as 
\begin{equation}
	\ell_{\mathrm{field}}=\frac{1}{H \times W}\sum_{j=1}^{H} \sum_{k=1}^{W} |\hat{T}_{j, k}-{T}_{j, k}|,
\end{equation}
where $H$ and $W$ denote the height and width of the temperature field image. 

Generally speaking, the ground-truth temperature distribution is continuous. However, the field loss may lead to a coarse reconstruction performance. To this end, this work further uses gradient loss (\cite{Nano}) as a regularization to improve the spatial smoothness of the reconstructed temperature field. The gradient loss provide the absolute error of gradient information between the reconstructed temperature field and the real temperature and is defined as
\begin{equation}
	\begin{aligned}
		\ell_{\mathrm{grad}} = & \frac{1}{H \times W} \left( \sum_{j=1}^{H} \sum_{k=1}^{W} \left\| |\hat{T}_{j, k}-\hat{T}_{j+1, k}|- |T_{j, k}-T_{j+1, k}| \right\| + \left\| |\hat{T}_{j, k}-\hat{T}_{j, k+1}|-|T_{j, k}-T_{j, k+1}| \right\| \right).
	\end{aligned}
\end{equation}

Then the total loss for the training of adaptive UNet can be formulated as a combination of field loss and gradient loss, where $\lambda$ is the regularization weight.
\begin{equation}
	\ell=\ell_{\mathrm{field}}+\lambda \cdot \ell_{\mathrm{grad}}.
\end{equation}

As for the training of MLP, the mean absolute error as a loss function can achieve a good performance. The loss function can be formulated as
\begin{equation}
	\ell^{'}=\frac{1}{n^{'}}\sum_{i=1}^{n^{'}} |\hat{T}_i-{T}_i|,
\end{equation}
where $n^{'}$ is the number of the target temperature points.

Given a set of temperature observation points $T_{ob}$ as input, the reconstructed temperature field $\hat T\in R^{200\times200}$ can be obtained as output from the deep learning neural networks. The training objective is to minimize the margin between the reconstructed temperature field $\hat T$, and the ground-truth simulated one $T$ so that the model can fit the provided data. Once the training of deep neural networks finishes, it can play as a surrogate model for reconstruction, which means reconstructing the corresponding temperature field w.r.t arbitrary temperature observation points.

\section{Experimental setups}
\label{sec:4}

\begin{figure*}[htbp]
	\centering
	\subfigure[Case 1, random layout]{
		\centering
		\includegraphics[width=0.485\linewidth]{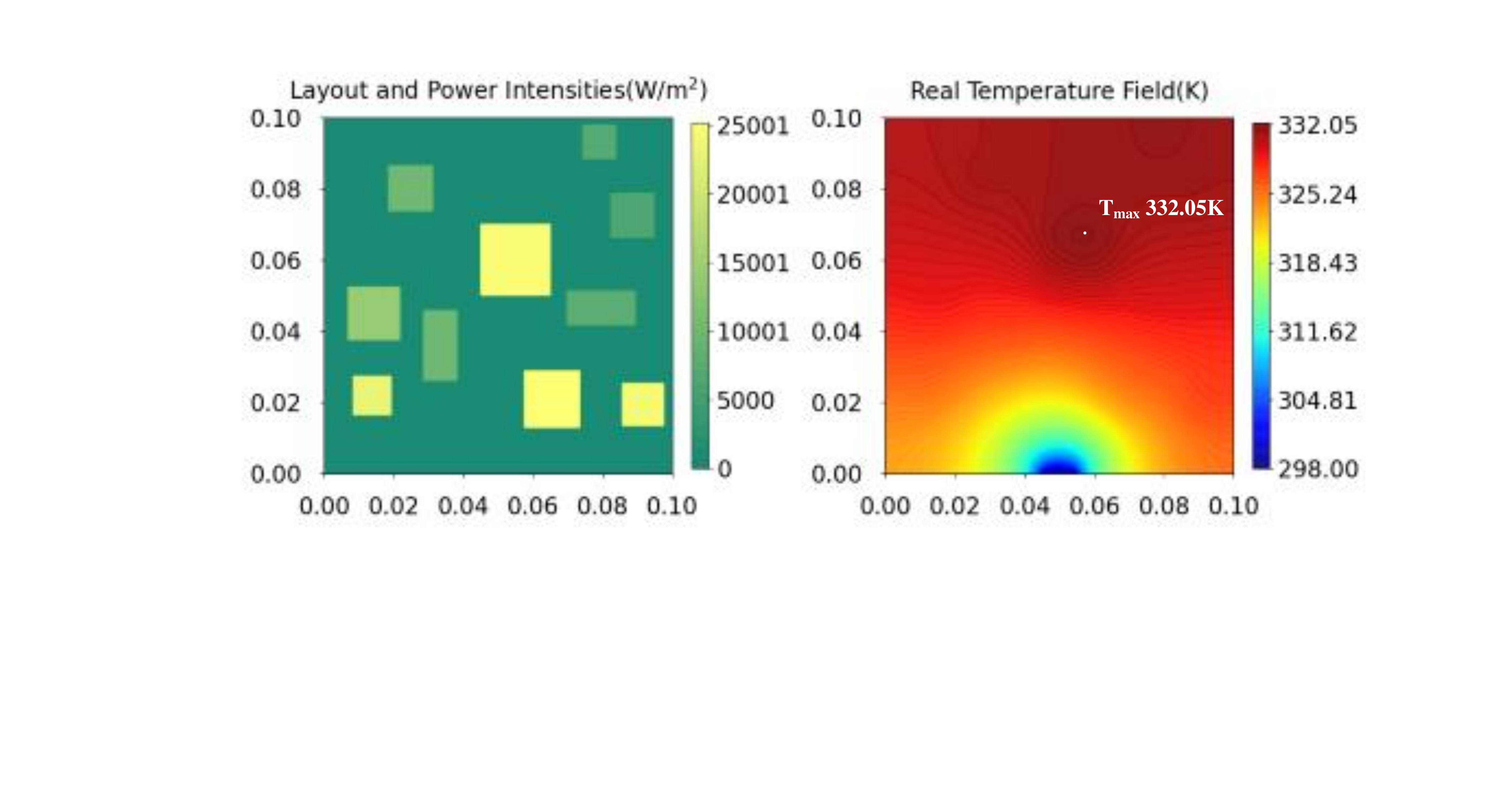}
		\label{fig:layout_case1}
	}
	\subfigure[Case 2, compact layout]{
		\centering
		\includegraphics[width=0.485\linewidth]{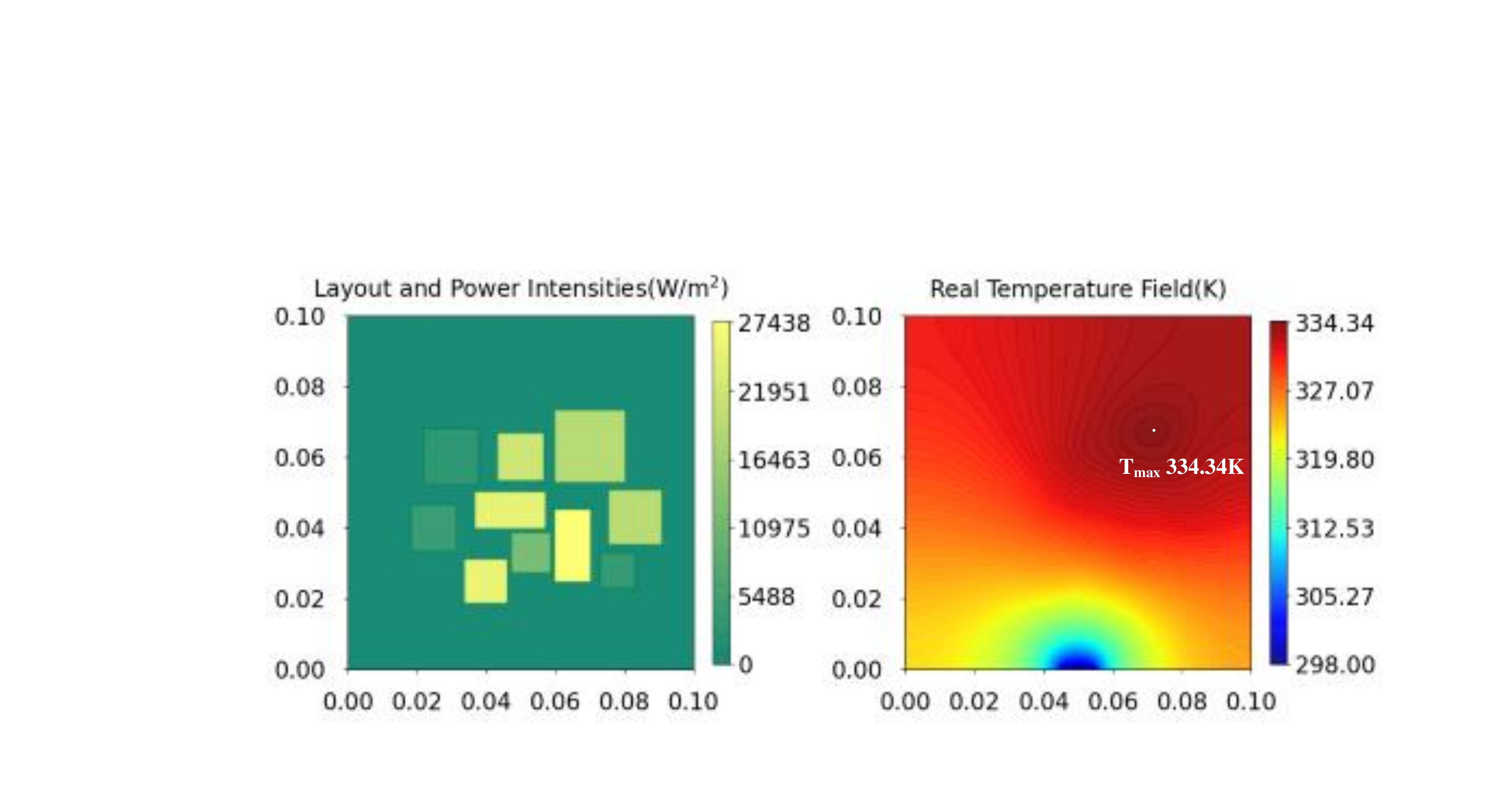}
		\label{fig:layout_case2}
	}
	\caption{Two cases with different layouts}
	\label{fig:layout}
\end{figure*}

In our experiments, 20k simulated samples are generated for the training process where $80\%$ is for training and $20\%$ for testing. To test the feasibilities of the surrogate model for handling different heat source layouts, Fig. \ref{fig:layout_case1} and fig. \ref{fig:layout_case2} present two typical cases which have a random heat source layout and a compact heat source layout and are denoted as Case 1 and Case 2, respectively.
\begin{figure*}[htb]
	\centering
	\subfigure[Case 1, special samples]{
		\includegraphics[width=0.8\linewidth]{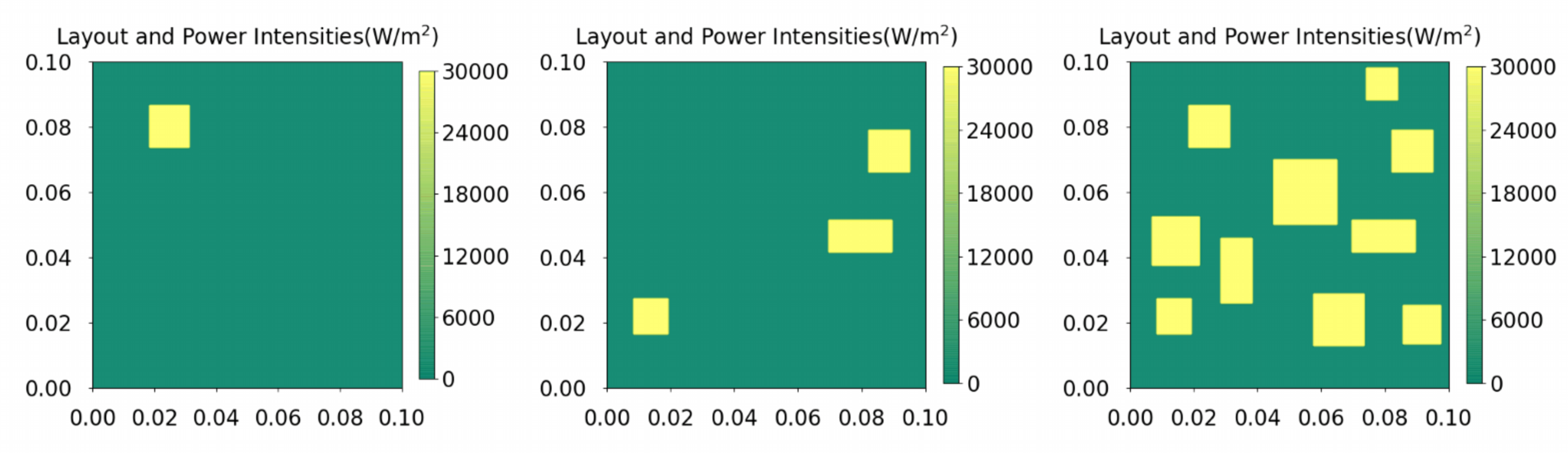}
		\label{fig:c1_sp}
	}
	\subfigure[Case 2, special samples]{
		\includegraphics[width=0.8\linewidth]{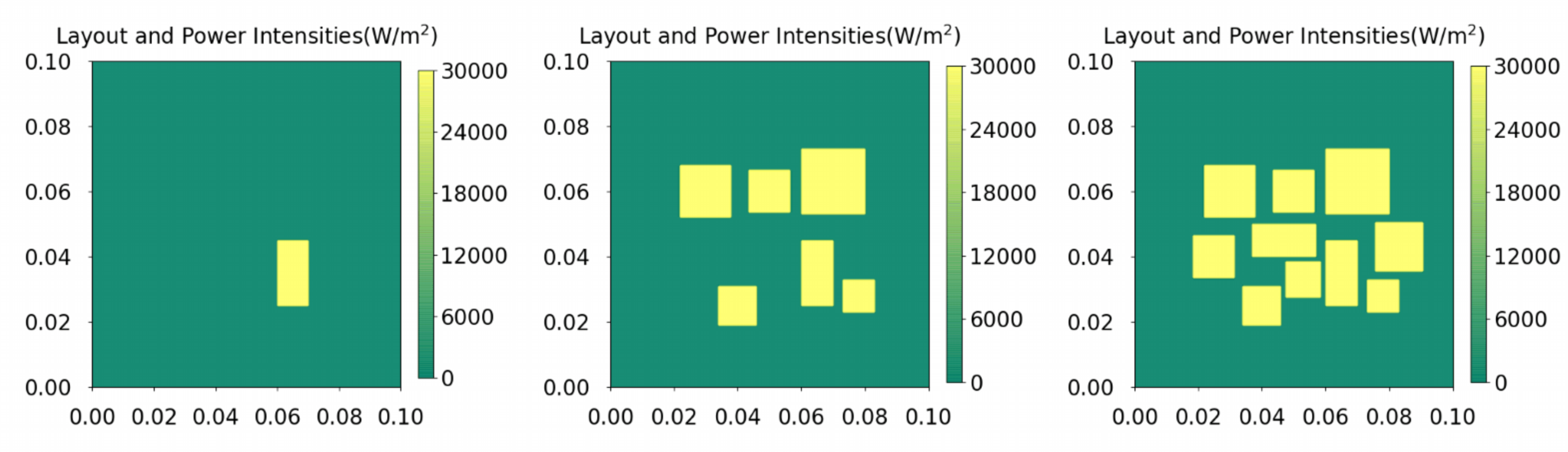}
		\label{fig:c2_sp}
	}
	\caption{Special test dataset with extreme power intensities}
	\label{fig:special_test}
\end{figure*}

Despite the diversity of general samples as Fig. \ref{fig:layout} shows, some special samples with extreme power intensities are hard to generate by the strategy of data generation. Hence, to validate the generalization ability of the proposed method, a special test dataset with the setting of extreme heat source power intensities, i.e., $0W/m^2$ or $30000W/m^2$, is generated for each case as Fig. \ref{fig:special_test} shows. For there are ten heat sources in the domain, each one has two extreme states, so the special test dataset is consists of $2^{10}=1024$ samples for each case.

The deep neural network is implemented using Pytorch. Adam (\cite{Adam}) is chosen as the optimizer method. The regularization weight, training epoch, initial learning rate, and batch size are set to 0.1, 200, 0.01, and 16, respectively. The structure of the shallow MLP is set to ’m-32-32-52', where $m$ is the number of the observation points as the input.

To evaluate the reconstruction performance of the proposed method comprehensively, four metrics about the temperature field and maximum temperature information we are concerned about are introduced here, namely the mean absolute error (MAE), the component-constrained mean absolute error (CMAE), the maximum absolute error (MaxAE), and the absolute error of the maximum temperature (MT-AE).

Mean absolute error (MAE), which means the mean value of absolute error between the reconstructed temperature matrix $\hat T$ and the ground-truth temperature matrix $T$, is formulated as
\begin{equation}
\textup{MAE}=\frac{1}{H \times W}\sum_{j=1}^{H} \sum_{k=1}^{W} |\hat{T}_{j, k}-{T}_{j, k}|,
\end{equation}
where $H$ and $W$ are the row and column of the output image. MAE will be used as the main evaluation metric in the following chapters. 

Component-constrained mean absolute error (CMAE), which means the mean absolute error of the area that heat sources cover, is formulated as
\begin{equation}
\textup{CMAE}=\frac{1}{|\Omega_{com}|} \sum_{(j, k \in \Omega_{com})} \left|\hat{T}_{j, k}-T_{j, k}\right|.
\end{equation}

Besides, the maximum temperature is always considered as an essential indicator of system condition. Therefore, we introduced two evaluation metrics concerned with maximum temperature.

Maximum absolute error (MaxAE), which means the maximum absolute error of the reconstructed temperature field, is formulated as
\begin{equation}
\textup{MaxAE}=\textup{max} \left|\hat{T}_{j,k}-T_{j,k}\right|.
\end{equation}

Absolute error of the maximum temperature (MT-AE), which means the maximum error between the reconstructed maximum temperature and the ground-truth maximum temperature (\cite{Benchmark}), is formulated as
\begin{equation}
\textup{MT-AE}=|\textup{max}(\hat{T}_{j,k})-\textup{max}(T_{j,k})|.
\end{equation}

\begin{figure}[htbp]
	\centering
	\subfigure[MAE]{
		\includegraphics[width=0.4\linewidth]{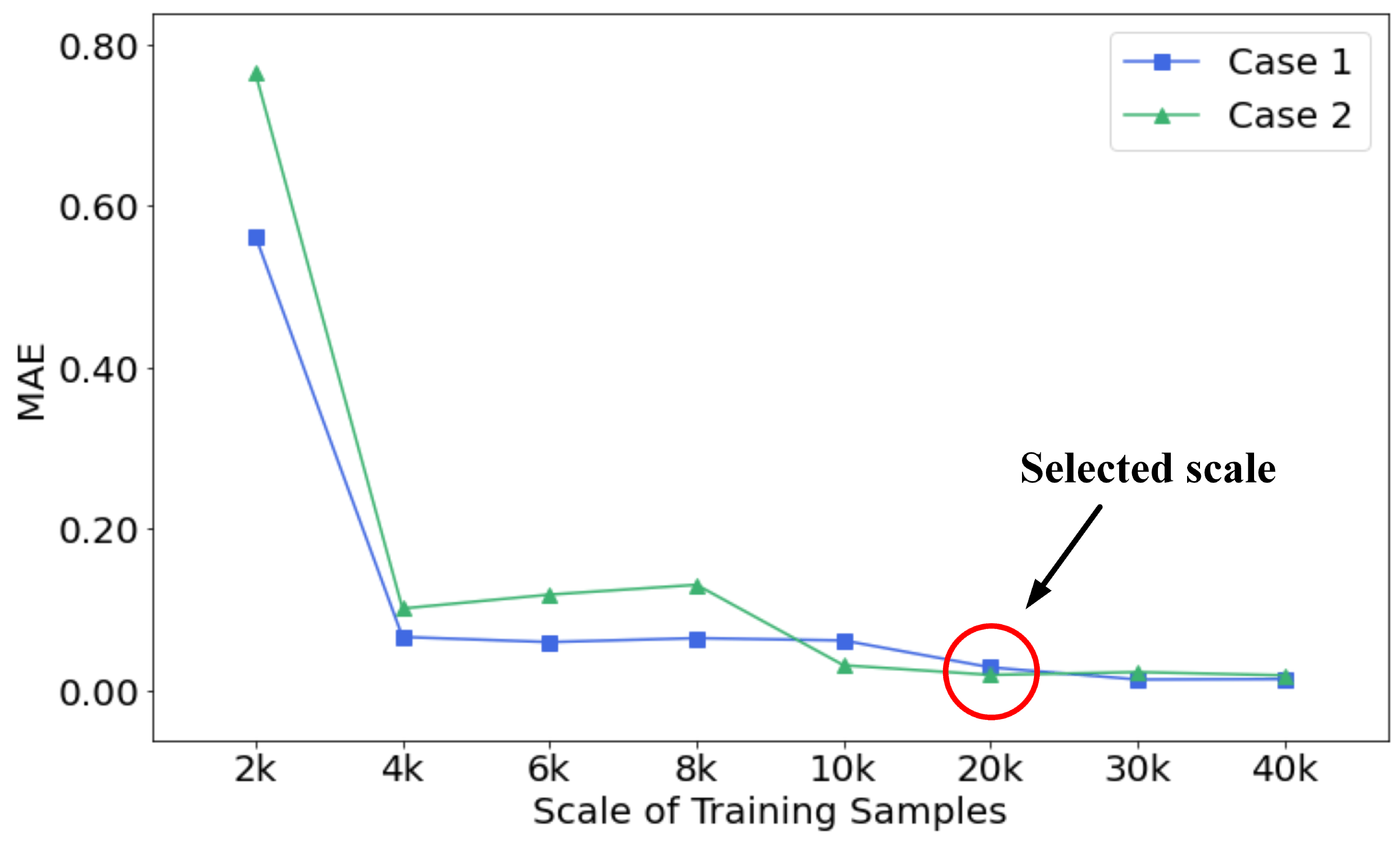}
	}
	\subfigure[CMAE]{
		\includegraphics[width=0.4\linewidth]{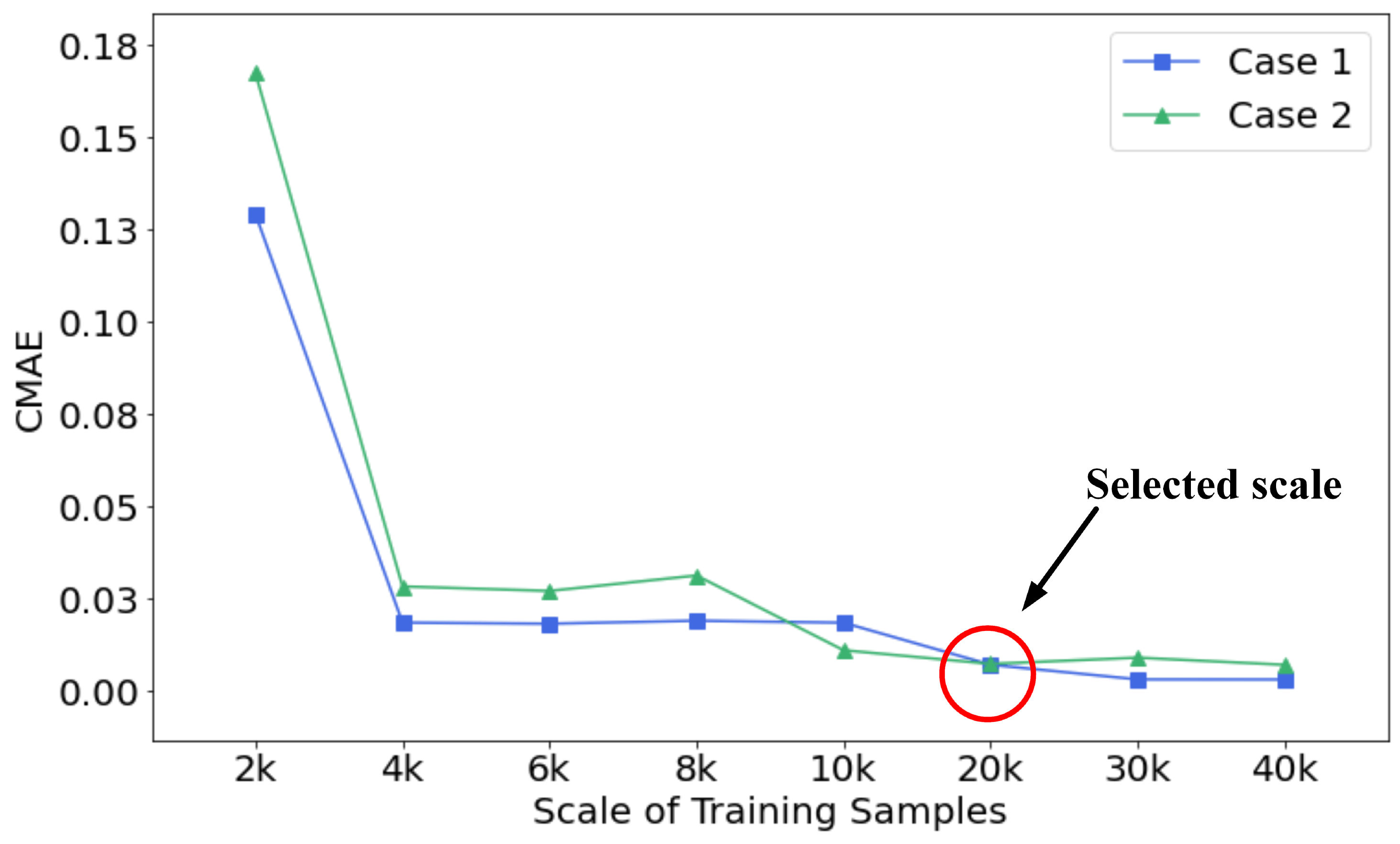}
	}
	\subfigure[MaxAE]{
		\includegraphics[width=0.4\linewidth]{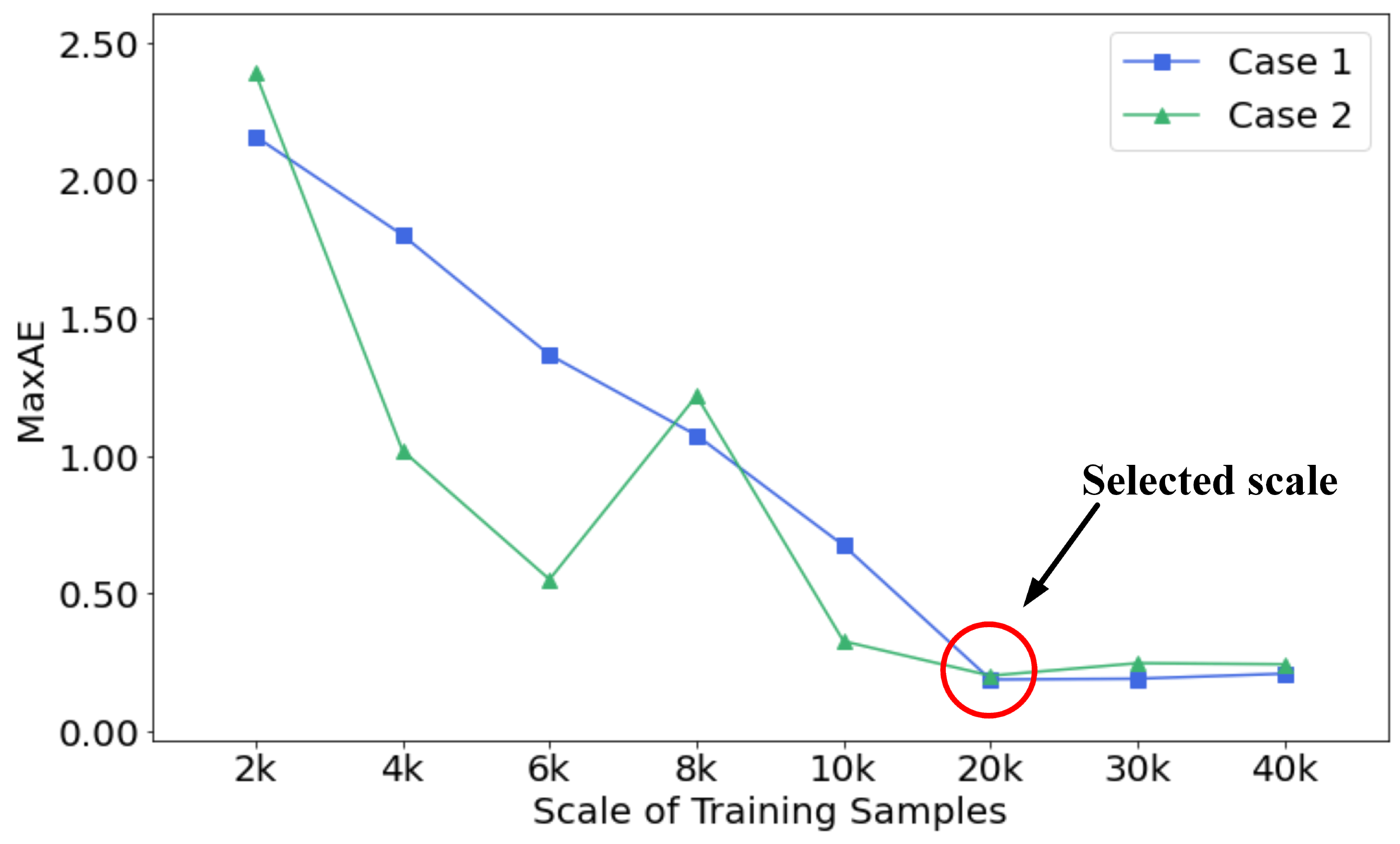}
	}
	\subfigure[MT-AE]{
		\includegraphics[width=0.4\linewidth]{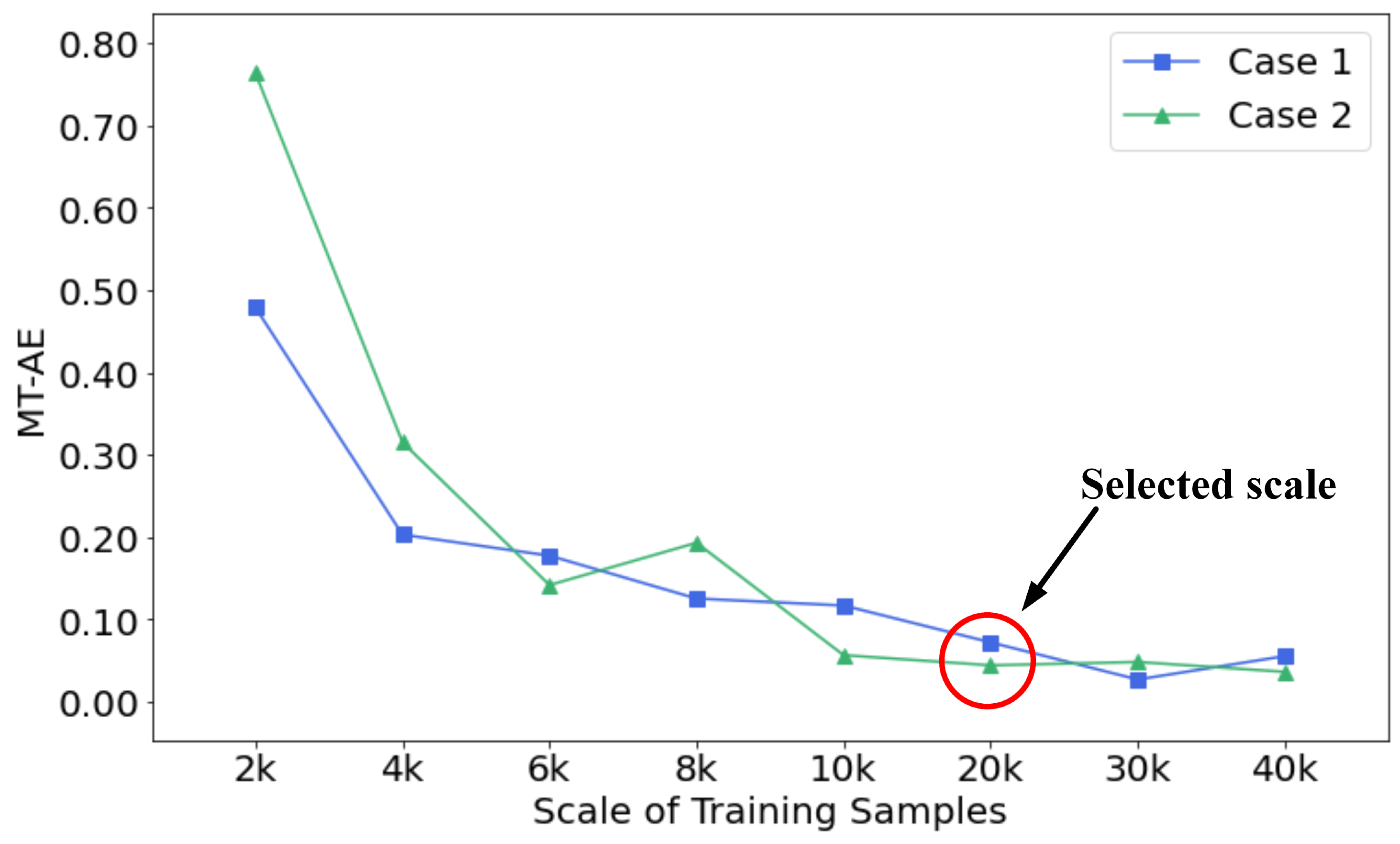}
	}	
	\caption{Performances with different scales of training samples. The temperature observation points are set as Fig. \ref{fig:different_layout_performance} shows.}
	\label{fig:training_scale}
\end{figure}

As a rule of thumb, the more training data, the better performance since overfitting can be alleviated by diverse data. To choose a proper scale of training samples, the deep neural networks are trained with different numbers of training samples, i.e., 2k, 4k, 6k, 8k, 10k, 20k, 30k, and 40k. For the sake of simplicity, the temperature observation points are deployed uniformly on the domain, and the number is set to 16 for this issue (see subsection \ref{sec:5.2.2} for more details). As shown in Fig. \ref{fig:training_scale}, the increase of the training samples leads to better reconstruction performance generally. More training samples always improve the completeness of data distribution until the data distribution is fully covered. Another critical point here is that when the number of training samples increases to a certain level, i.e., 20K, less improvement is observed. The reason is that the training data have almost covered the data distribution. Thus little rise to the performance can be acquired by adding more data singly. For simplicity, 20k is chosen as the size of the dataset in the following.

\section{Results and discussion}
\label{sec:5}

In this section, two cases with varying degrees of complexity are studied to demonstrate the excellent performance of the deep learning-based surrogate model based on patchwise training. Firstly, the configurations of two cases, training hyperparameters and evaluation metrics, are presented. Then the reconstruction performances via patchwise training approach on general and special datasets are discussed. To validate the reconstruction accuracy and generalization of the proposed methods, the performances for the special dataset, for two cases with different observation numbers and locations are investigated. At last, the performance of the proposed method is compared with that of traditional surrogate model methods for the precision of TFR.

\subsection{Reconstruction performance analysis}
\label{sec:5.1}

In this section, firstly, the effectiveness and feasibilities of the proposed deep learning method for the TFR problem are presented. Then, the issue of large temperature gradient and the performance improvement via patchwise training approach are discussed comprehensively.

\subsubsection{Performance with different heat source layouts}
\label{sec:5.1.1}

\begin{figure*}[htbp]
	\centering
	\subfigure[Case 1, random layout]{
		\centering
		\includegraphics[width=1\linewidth]{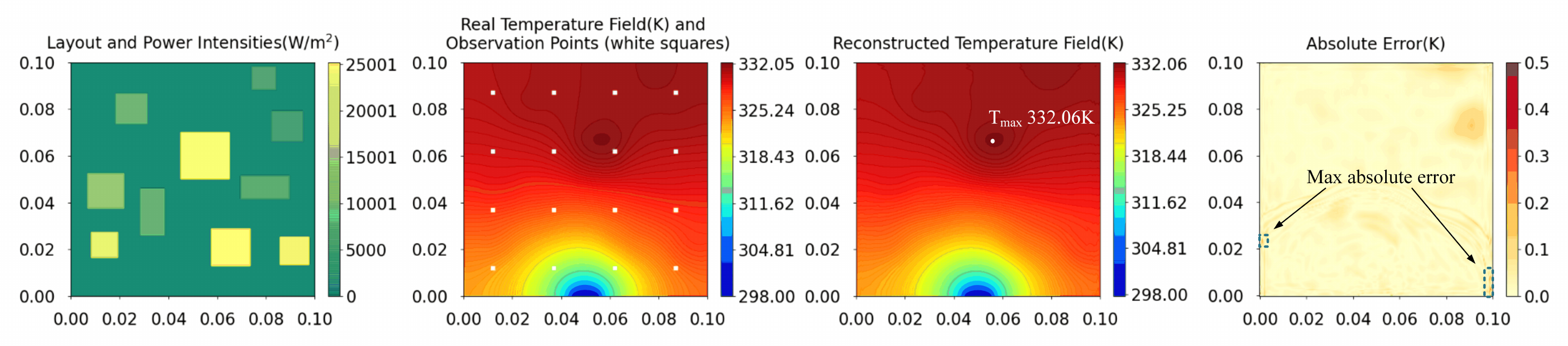}
		\label{fig:case1}
	}
	\subfigure[Case 2, compact layout]{
		\centering
		\includegraphics[width=1\linewidth]{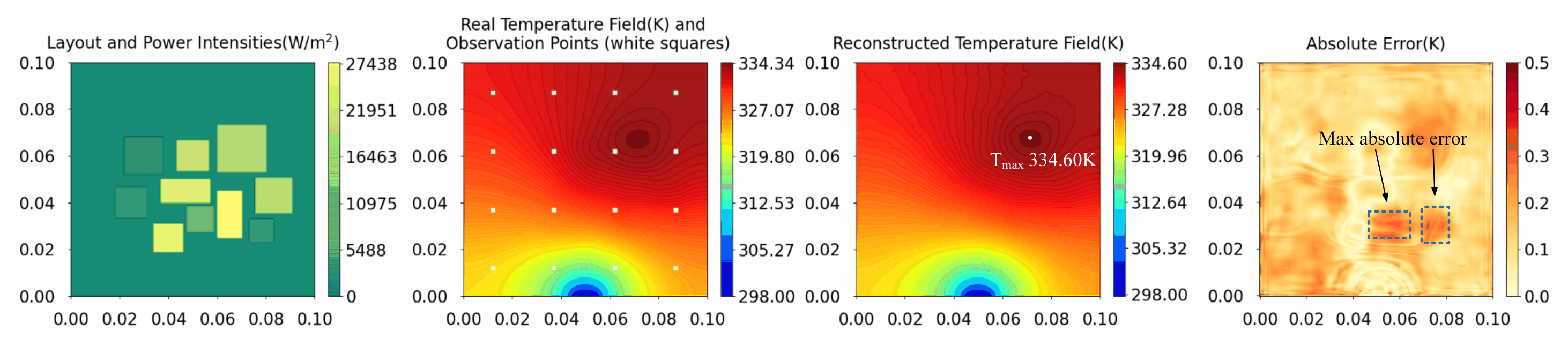}
		\label{fig:case2}
	}
	\caption{The performances of two different heat source layouts.}
	\label{fig:different_layout_performance}
\end{figure*}

For different heat source layouts with $4\times4$ observation points, the reconstruction performances are listed in Table \ref{tab:1} and several reconstructed temperature field examples are provided in Fig. \ref{fig:different_layout_performance}. As the results show, the MAE, CMAE, and MT-AE of both cases are less than 0.03K, 0.008K and 0.2K, respectively. As for MaxAE, which evaluates the maximum absolute error, is smaller than 1K. It means the error between the reconstruction field and ground truth is small and uniform. Since the interaction of heat sources is more serious for the compact layout, case 2 represents a much more complicated situation. All evaluation metrics of case 2 are just slightly larger. Significantly, the reconstructed temperature distributions, e.g., the peaks and valleys, are in good agreement with the ground-truth ones. These outstanding performances demonstrate that the deep learning-based surrogate model can reconstruct the temperature field accurately from limited observation. 

\begin{table*}[htbp]
	\caption{The performances of two cases with different heat source layouts and MAE, CMAE, MaxAE, and MT-AE, are used as the evaluation metrics. NO denotes the number of observation points in this paper.}
	\label{tab:1}
	\centering
	\begin{tabular}{c c c c c c c}
		\hline\noalign{\smallskip}
		& Test set & NO & MAE & CMAE & MaxAE & MT-AE  \\
		\noalign{\smallskip}\hline\noalign{\smallskip}
		\multirow{2}{*}{Case 1}\centering & general test & $4\times4$ & 0.0216 & 0.0057 & 0.2614 & 0.0790 \\
		& special test & $4\times4$ & 0.1648 & 0.0355 & 0.6454 & 0.2383 \\
		\hline\noalign{\smallskip}
		\multirow{2}{*}{Case 2}\centering &general test & $4\times4$ & 0.0246 & 0.0077 & 0.8362 & 0.1360 \\
		& special test & $4\times4$ & 0.1559 & 0.0394 & 0.9452 & 0.1880 \\
		\hline\noalign{\smallskip}
	\end{tabular}
\end{table*}

\subsubsection{Performance imrovement through patchwise training approach}
\label{sec:5.1.2}

\begin{figure}[htb]
	\centering
	\includegraphics[width=0.6\linewidth]{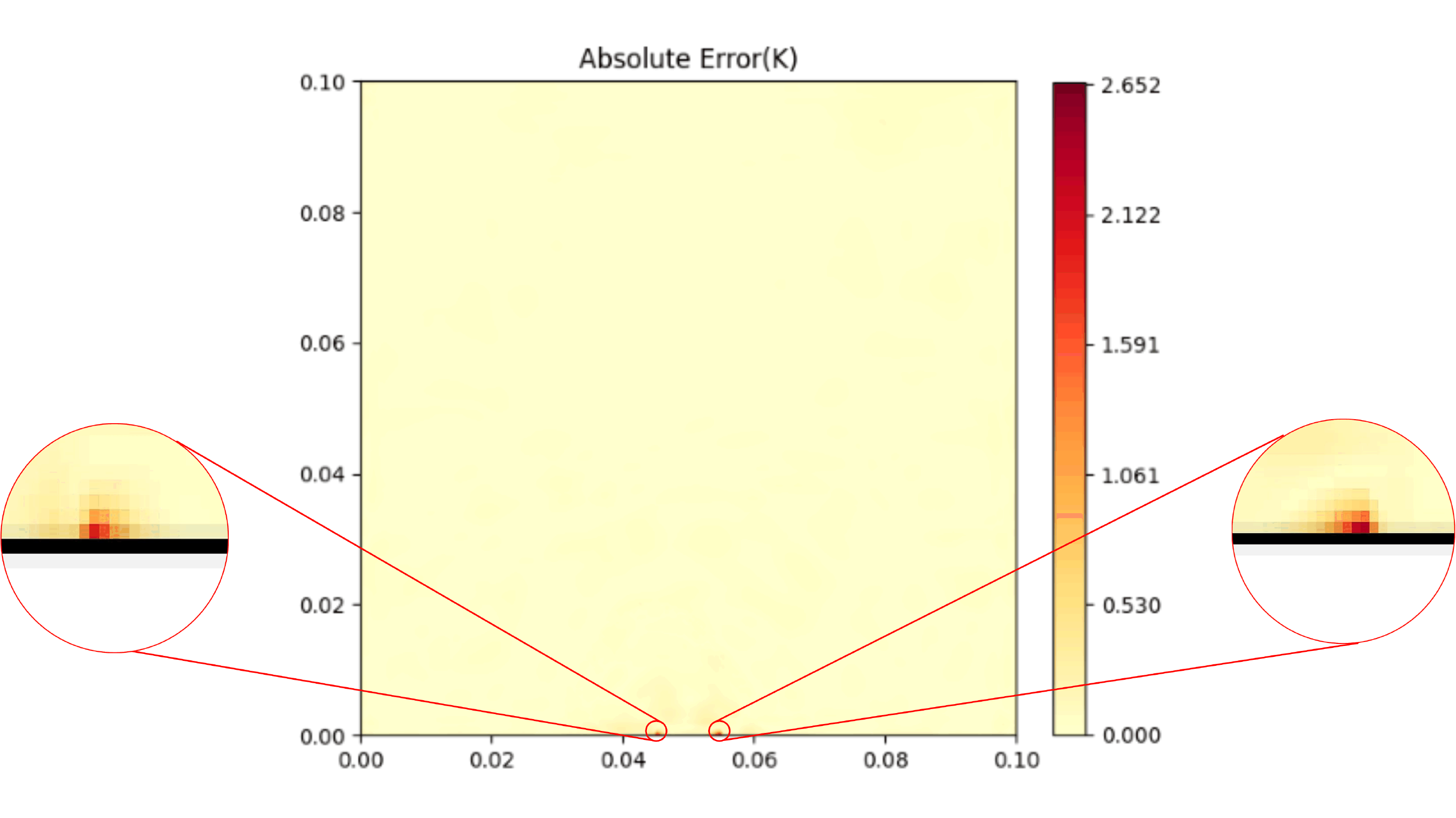}
	\caption{The detail of MaxAE in the begining and end of the heat sink.}
	\label{fig:heat_sink}
\end{figure}

\begin{figure}[htb]
	\centering
	\includegraphics[width=0.4\linewidth]{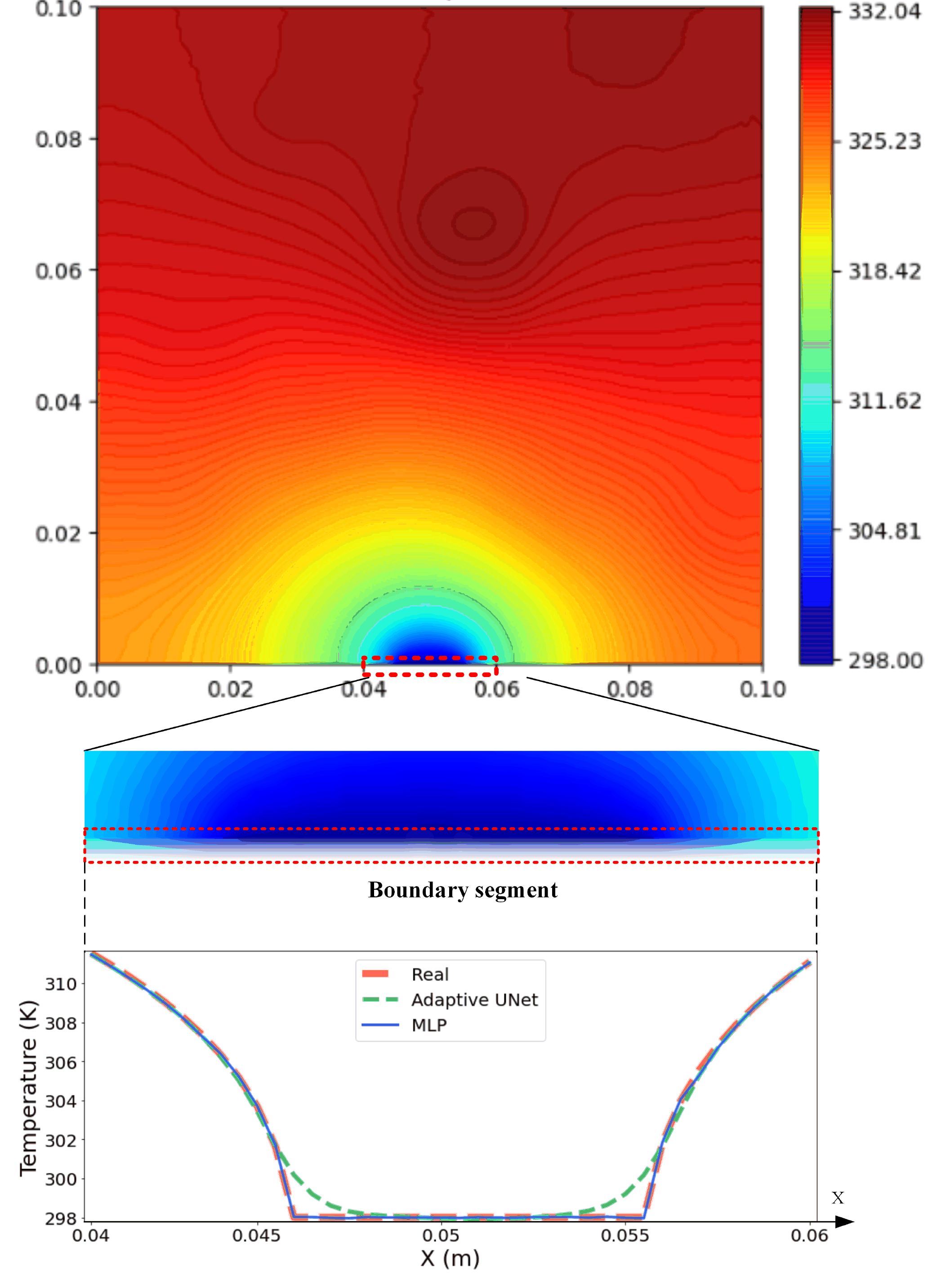}
	\caption{The temperature distribution along the bottom boundary (middle part) reconstructed by different methods..}
	\label{fig:heatsink_mlp}
\end{figure}

An interesting and important finding is that the MaxAE is mainly located in the domain near the heat sink where steep temperature gradients exist when only adaptive UNet is utilized to reconstruct the temperature field, as shown in Fig. \ref{fig:heat_sink}. For all other areas, the pixel error is less than 1K. This may be attributed to the convolutional neural network's characteristics that it is hard to learn and predict discontinuous distribution. Therefore, a shallow MLP is designed to build the direct mapping from the observation to the temperature of the small special domain as shown in Fig. \ref{fig:MLP}. As a result, the MLP performs well with this issue and improves the MaxAE metric obviously, as Table \ref{tab:2} shows. The results show that the MaxAE is reduced by 83.7$\%$, 47.8$\%$ for the general datasets, 68.8$\%$, 57.5$\%$ for the special datasets of Case 1 and Case 2, respectively. For the sake of illustration, the temperature distribution along the bottom boundary, where the x ranges from 0.04m to 0.06m, reconstructed by different methods is presented in Fig. \ref{fig:heatsink_mlp}. It is noticeable that the real temperature distribution is discontinuous at the beginning and end of the heatsink. Whereas the temperature distribution reconstructed by adaptive UNet is always continuous thus can not fit the real temperature well. For such a situation, the shallow MLP perfectly learns and predicts the discontinuous temperature distribution in the domain near the heatsink. 
\begin{table*}[htbp]
	\caption{The MaxAE improved by the MLP.}
	\label{tab:2}
	\centering
	\begin{tabular}{c c c c c c}
		\hline\noalign{\smallskip}
		\multirow{2}{*}{} & \multicolumn{2}{c}{Case 1} & \multicolumn{2}{c}{Case 2} \\
		& general set  & special set & general set  & special set \\
		\noalign{\smallskip}\hline\noalign{\smallskip}
		Adaptive UNet  & 1.6000 & 2.0676 & 1.6030 & 2.2268 \\
		\hline\noalign{\smallskip}
		Adaptive UNet + MLP  & 0.2614 & 0.6454 & 0.8362 & 0.9452 \\
		\hline\noalign{\smallskip}
		Improvement  & 83.7$\%$ & 68.8$\%$ & 47.8$\%$ & 57.5$\%$ \\
		\hline\noalign{\smallskip}   
	\end{tabular}
\end{table*}

\subsection{Generalization ability evaluation}
\label{sec:5.2}

To evaluate the generalization ability of the proposed method, the reconstruction performances for the special dataset, for two cases with different observation numbers and locations are investigated in this section.

\subsubsection{Performance for the special test dataset}
\label{sec:5.2.1}

\begin{figure*}[h]
	\centering
	\subfigure[Case 1, example of special test dataset]{
		\includegraphics[width=1\linewidth]{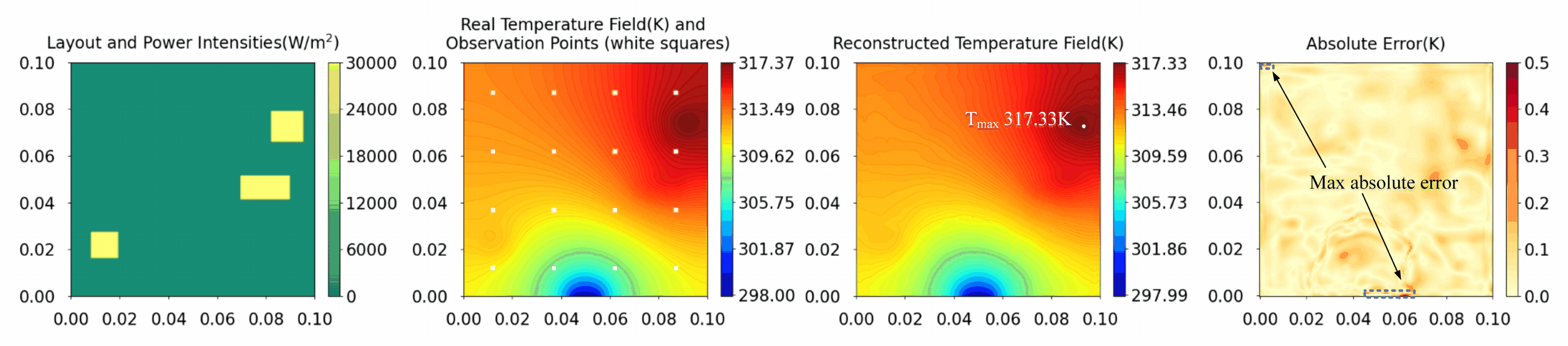}
		\label{fig:c1_sp_test_1}
	}
	\subfigure[Case 2, example of special test dataset]{
		\includegraphics[width=1\linewidth]{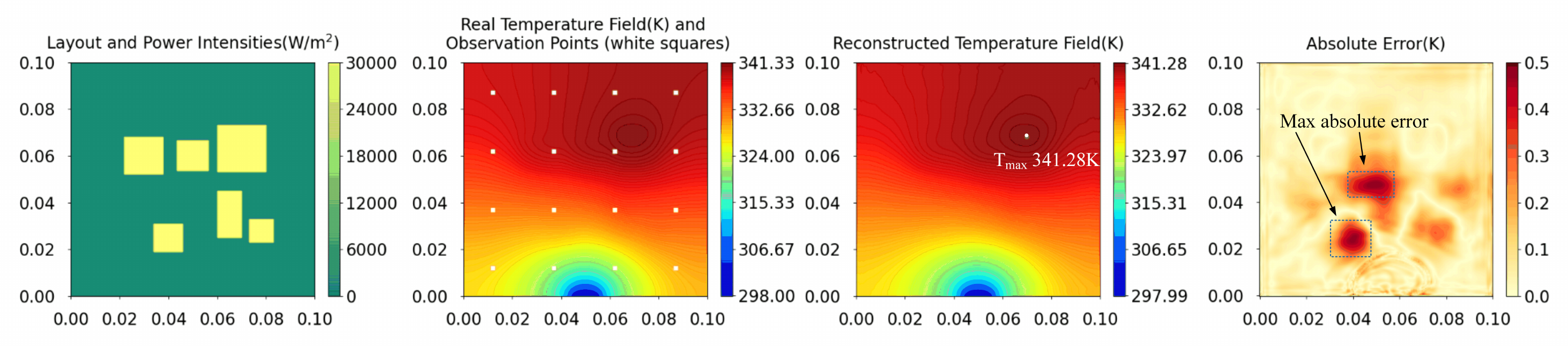}
		\label{fig:c2_sp_test_3}
	}
	\caption{The performance for special test dataset}
	\label{fig:special_test_performance}
\end{figure*}

Since each electronic equipment may be in a full power operation or power-off state, the power intensity has two extreme values, i.e., $0W/m^2$ and $30000W/m^2$. To reconstruct the temperature field with extreme power intensities accurately has prime importance. Therefore, a special test dataset as the subsection \ref{sec:4} described is used to validate the generalization ability of the proposed method. The reconstruction performances are listed in Table \ref{tab:1}. The MAE and CMAE of both cases are less than 0.2K and 0.04K, respectively, which is still a good performance. The MaxAE and MT-AE of both cases increase remarkably comparing with the general datasets. Two examples randomly selected from the special test samples and the corresponding reconstructed temperature fields are presented in Fig. \ref{fig:special_test_performance}. It can be observed that the performances are as good as that in Fig. \ref{fig:case1} intuitively, which suggests that the proposed method can work well for unseen samples that are not similar to the training data.

\subsubsection{Performance for different numbers of observation points}
\label{sec:5.2.2}

\begin{figure}[htbp]
	\centering
	\subfigure[MAE]{
		\includegraphics[width=0.48\linewidth]{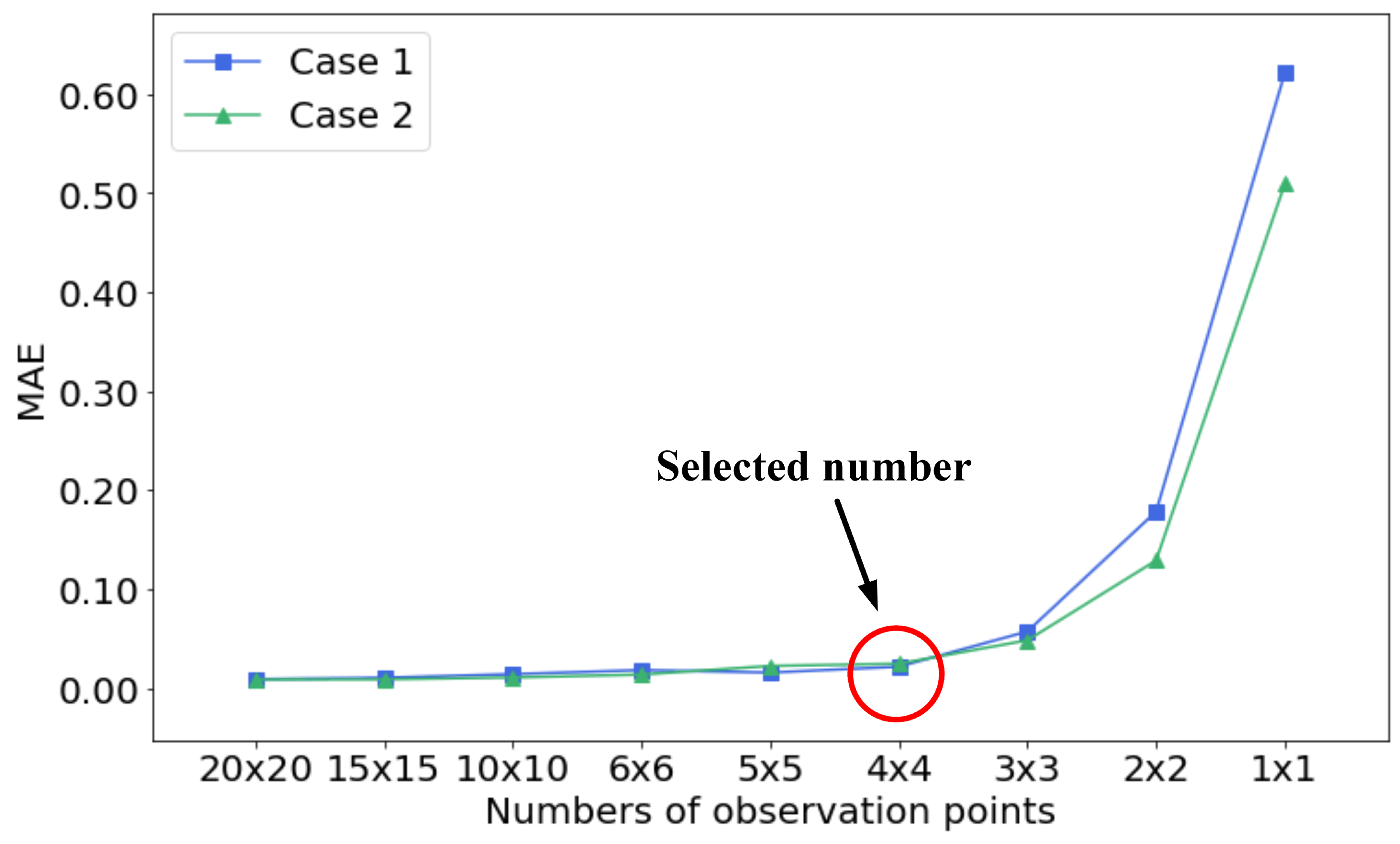}
		\label{fig:Numbers_MAE}
	}
	\subfigure[CMAE]{
		\includegraphics[width=0.48\linewidth]{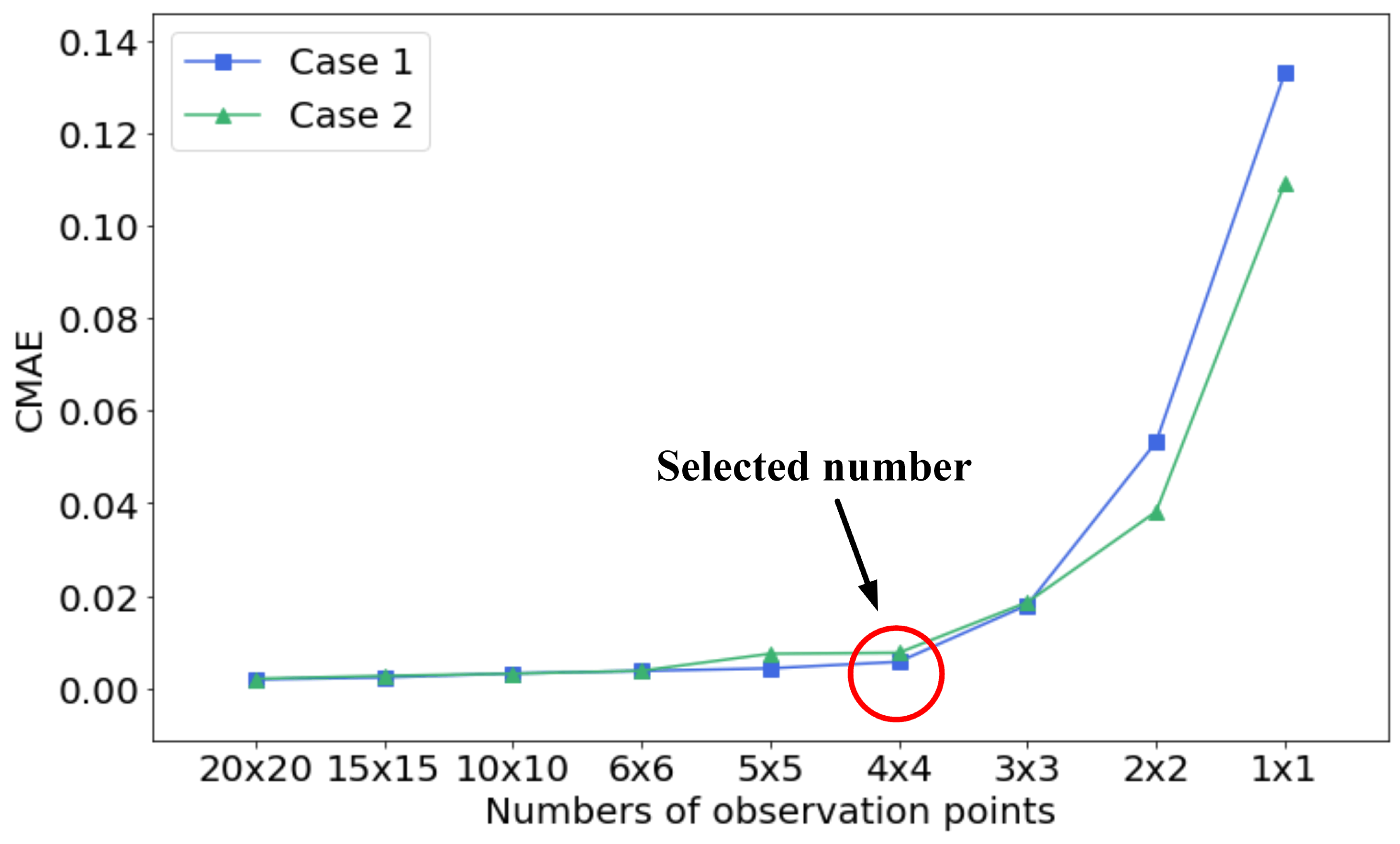}
		\label{fig:Numbers_CMAE}
	}
	\subfigure[MaxAE]{
		\includegraphics[width=0.48\linewidth]{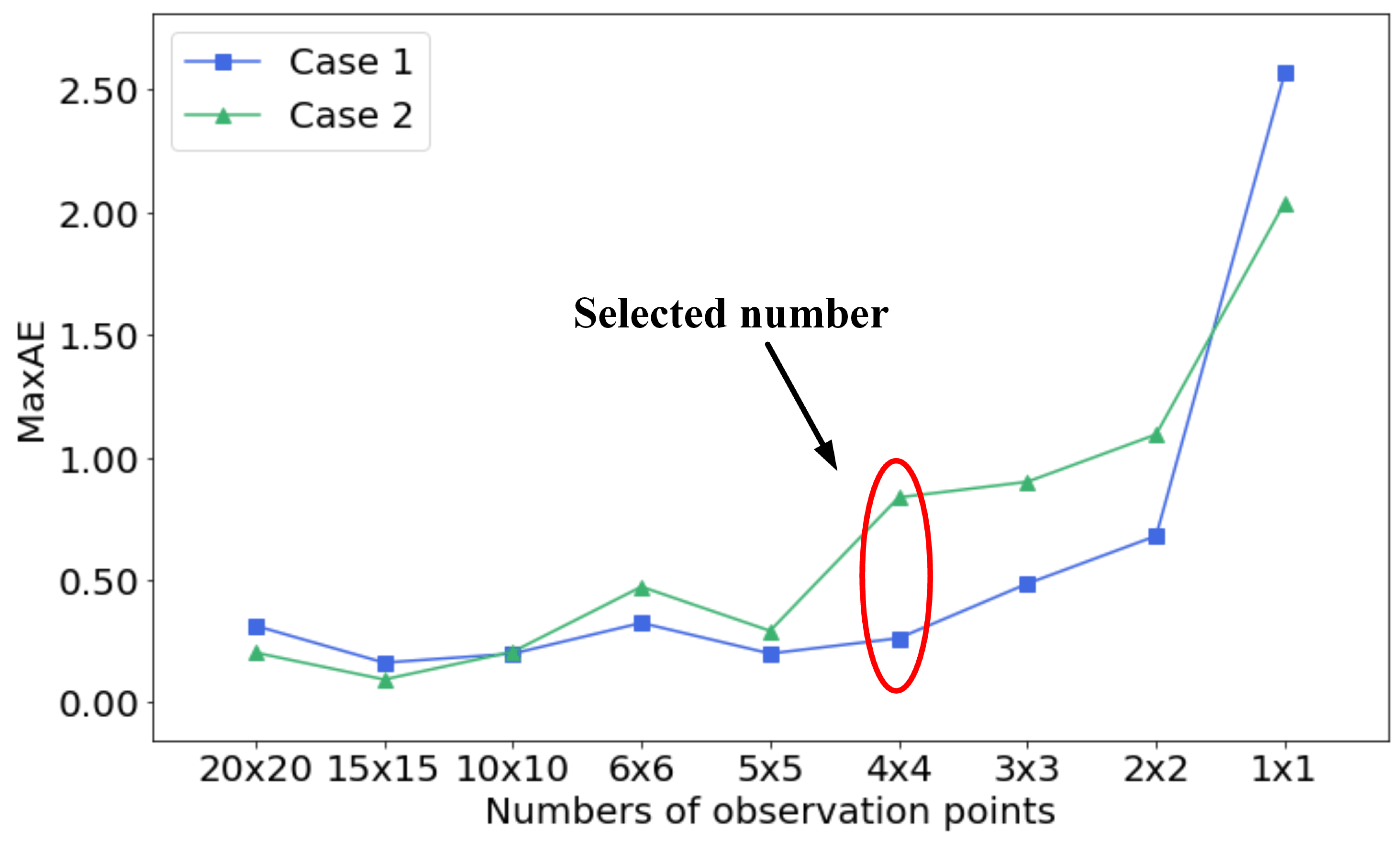}
		\label{fig:Numbers_MaxAE}
	}
	\subfigure[MT-AE]{
		\includegraphics[width=0.48\linewidth]{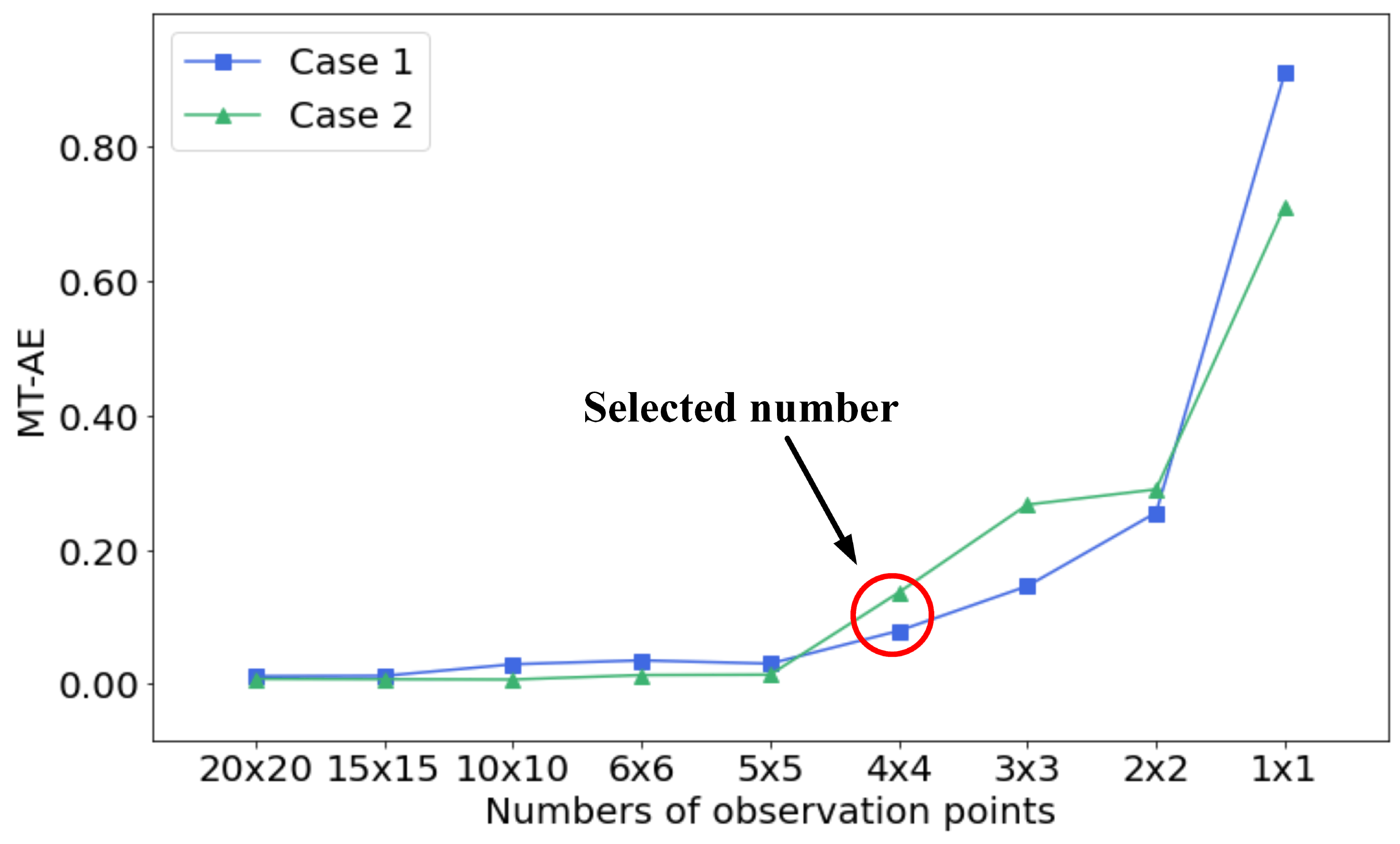}
		\label{fig:Numbers_MTAE}
	}
	\caption{Performance with different number of observation points}
	\label{fig:Numbers_of_observation_points}
\end{figure}

From the viewpoint of inverse problems, more temperature observation points usually contain more information, which is helpful to reconstruct the temperature field precisely. However, the measurement resource is always limited in real-world applications, and it pays to reduce the number of temperature observation points. To investigate the performance with different numbers of temperature observation points, six cases are introduced to conduct the training and reconstruction. As Fig. \ref{fig:Numbers_of_observation_points} shows, the reconstruction performance with more observation points is obviously better than that with fewer ones. Especially, a noticeable decline in reconstruction performance is observed until the number of observation points drops to $3\times3$ and blow. It shows that the proposed method can reconstruct temperature fields from very few observation while keeping high precision. Furthermore, the reconstructed temperature fields are almost identical under different input when the number of temperature observation points drops to $2\times2$ and blow. This means that too little information is not enough to uniquely determine the reconstruction temperature field. Considering the tradeoff between reconstruction precision and computation cost, $16$ temperature observation points are adopted in this work.

\subsubsection{Performance for different locations of observation points}
\label{sec:5.2.3}

\begin{figure*}[htbp]
	\centering
	\subfigure[Uniform selection strategy]{
		\centering
		\includegraphics[width=0.30\linewidth]{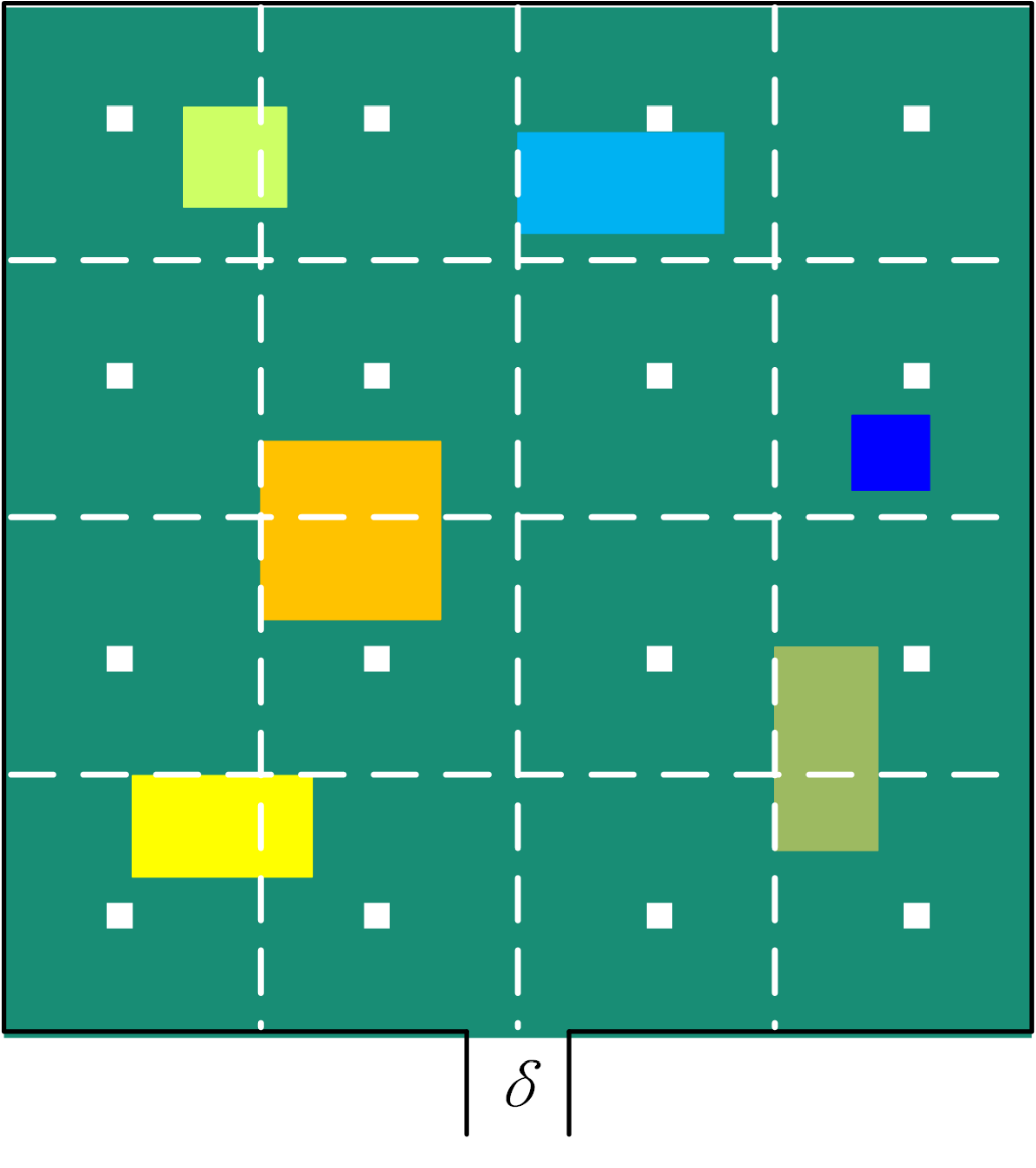}
		\label{fig:Three_strategies_a}
	}
	\subfigure[Random selection strategy]{
		\centering
		\includegraphics[width=0.30\linewidth]{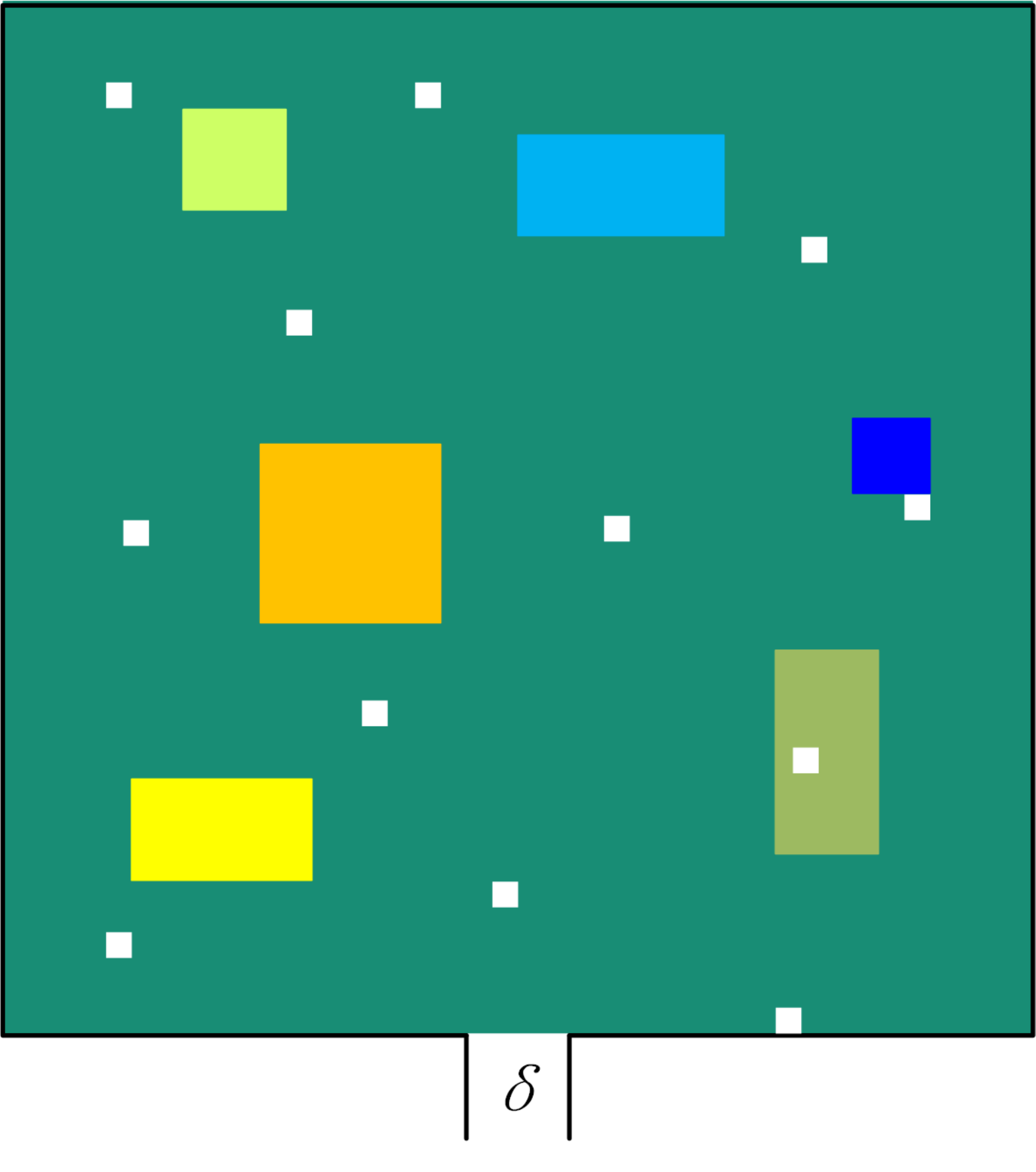}
		\label{fig:Three_strategies_b}
	}
	\subfigure[Physics-informed selection strategy]{
		\centering
		\includegraphics[width=0.30\linewidth]{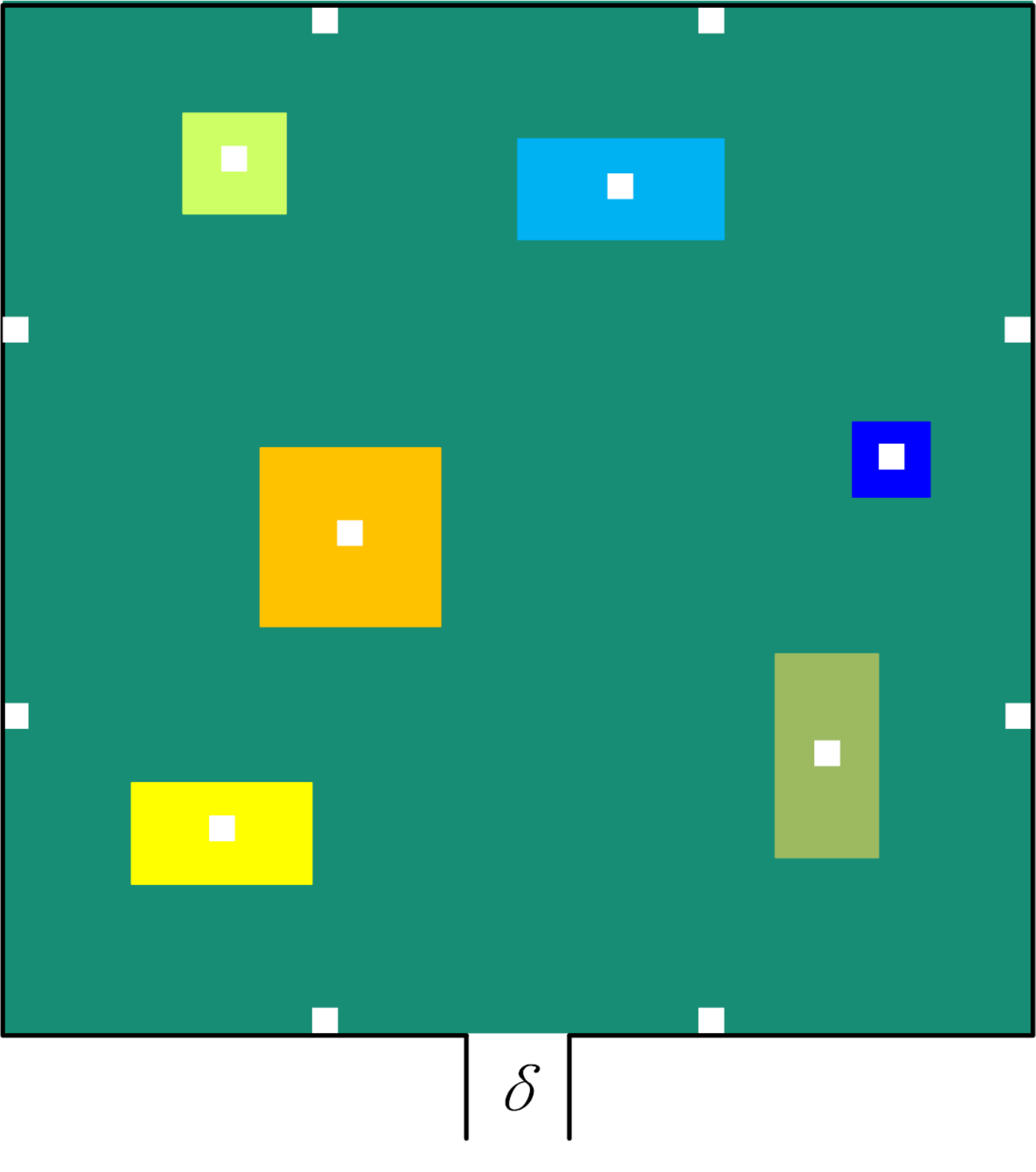}
		\label{fig:Three_strategies_c}
	}
	\caption{Three observation points locations selection strategies}
	\label{fig:Three_strategies}
\end{figure*}

To examine the performance with different temperature observation points locations of the proposed method, three selection strategies are employed to obtain different observation points locations. Given the number of observation points, i.e., 16, locations selection becomes critically important. Three strategies, including Uniform selection, Random selection, and Physics-informed selection, are considered to select appropriate observation locations as Fig. \ref {fig:Three_strategies} shows. Temperature observation points and components are denoted by small red squares and colorful rectangles. 16 observation points are placed in the domain. For the Uniform selection strategy, the domain is divided into 16 uniform blocks. Then the center of each block is chosen as the location of the observation point as Fig. \ref {fig:Three_strategies_a} shows. For the Random selection strategy, 16 randomized positions are generated by the program as the locations of observation points as Fig. \ref {fig:Three_strategies_b} shows. The distribution of observation points is not uniform. Namely, some points are too dense while the others are too sparse. There is no doubt that the components' temperature is what we care about. On the boundaries, temperature gradients are often great, especially in the adjacent area of the heat sink. Thus, for the Physics-informed selection strategy, the centers of the components are first chosen as the locations of observation points. Then the remaining points are placed on the boundaries as Fig. \ref {fig:Three_strategies_c} shows. 

\begin{table*}[htbp]
	\caption{The performance with different locations of observation points and corresponding results.}
	\label{tab:3}
	\centering
	\begin{tabular}{ c c c c  c c c}
		\hline\noalign{\smallskip}
		& strategy & NO & MAE & CMAE & MaxAE & MT-AE \\
		\noalign{\smallskip}\hline\noalign{\smallskip}
		\multirow{4}{*}{Case 1}\centering & Uniform selection & $4\times4$ & 0.0216 & 0.0057 & 0.2614 & 0.0790 \\
		& Random selection 1         & $16$ & 0.0758 & 0.0231 & 0.5031 & 0.1733 \\
		& Random selection 2         & $16$ & 0.1553 & 0.0344 & 0.8511 & 0.2757 \\
		& Physics-informed selection & $16$ & 0.0113 & 0.0020 & 0.2602 & 0.0160 \\
		\hline\noalign{\smallskip}
		\multirow{4}{*}{Case 1}\centering & Uniform selection  & $4\times4$ & 0.0246 & 0.0077 & 0.8362 & 0.1360 \\
		& Random selection 1         & $16$ & 0.0347 & 0.0148 & 0.4584 & 0.0247 \\
		& Random selection 2         & $16$ & 0.0301 & 0.0101 & 0.3379 & 0.0345 \\
		& Physics-informed selection & $16$ & 0.0147 & 0.0030 & 0.4008 & 0.0358 \\
		\hline\noalign{\smallskip}
	\end{tabular} 
\end{table*}

\begin{figure*}[htbp]
	\centering
	\subfigure[Uniform selection, Case 1]{
		\includegraphics[width=1\linewidth]{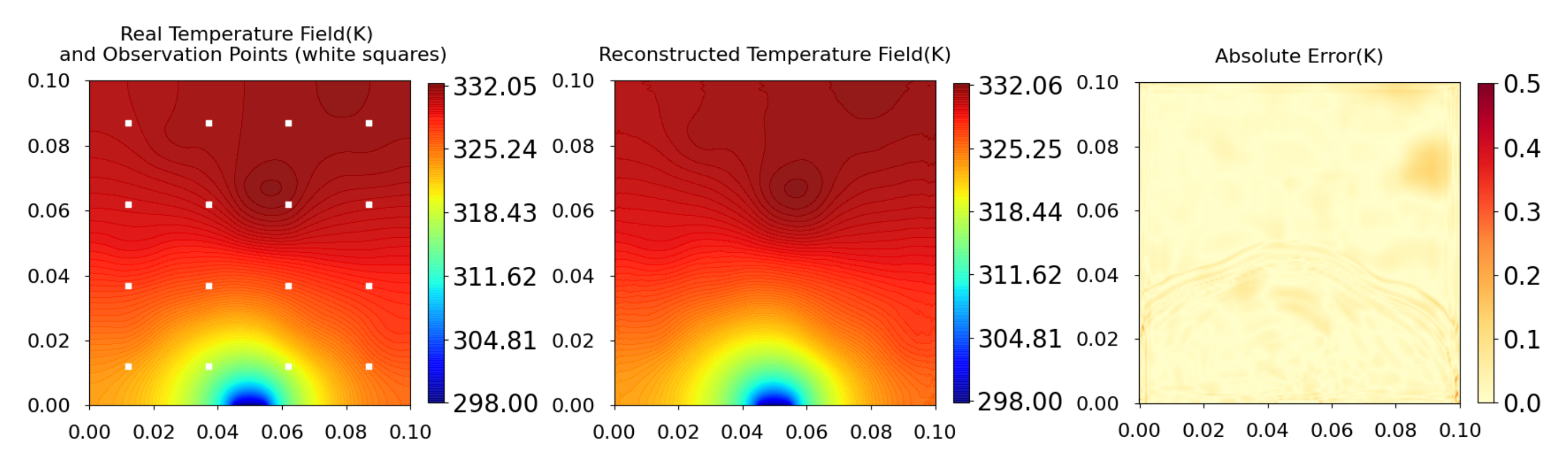}
		\label{fig:us_c1}
	}
	\subfigure[Random selection 1, Case 1]{
		\includegraphics[width=1\linewidth]{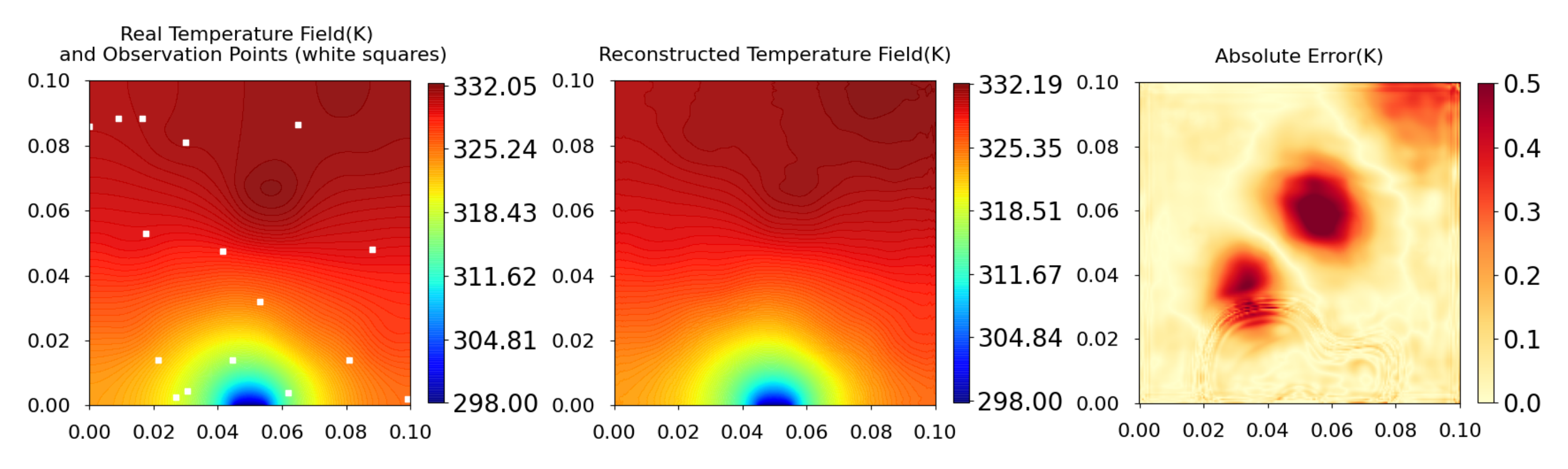}
		\label{fig:rs1_c1}
	}
	\subfigure[Random selection 2, Case 1]{
		\includegraphics[width=1\linewidth]{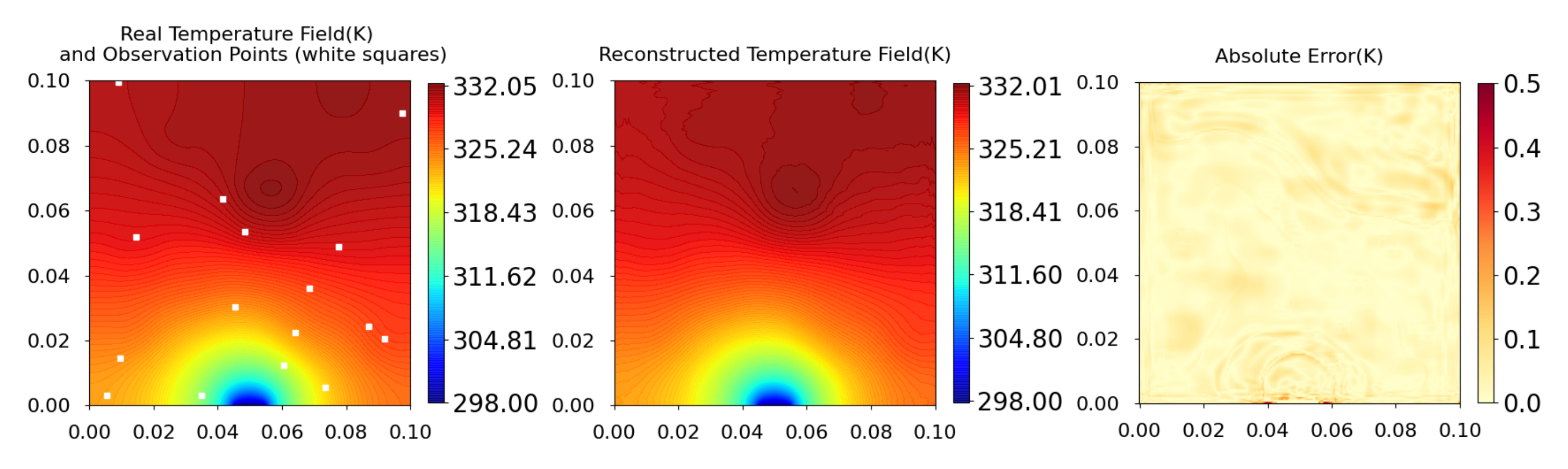}
		\label{fig:rs2_c1}
	}
	\subfigure[Physics-informed selection, Case 1]{
		\includegraphics[width=1\linewidth]{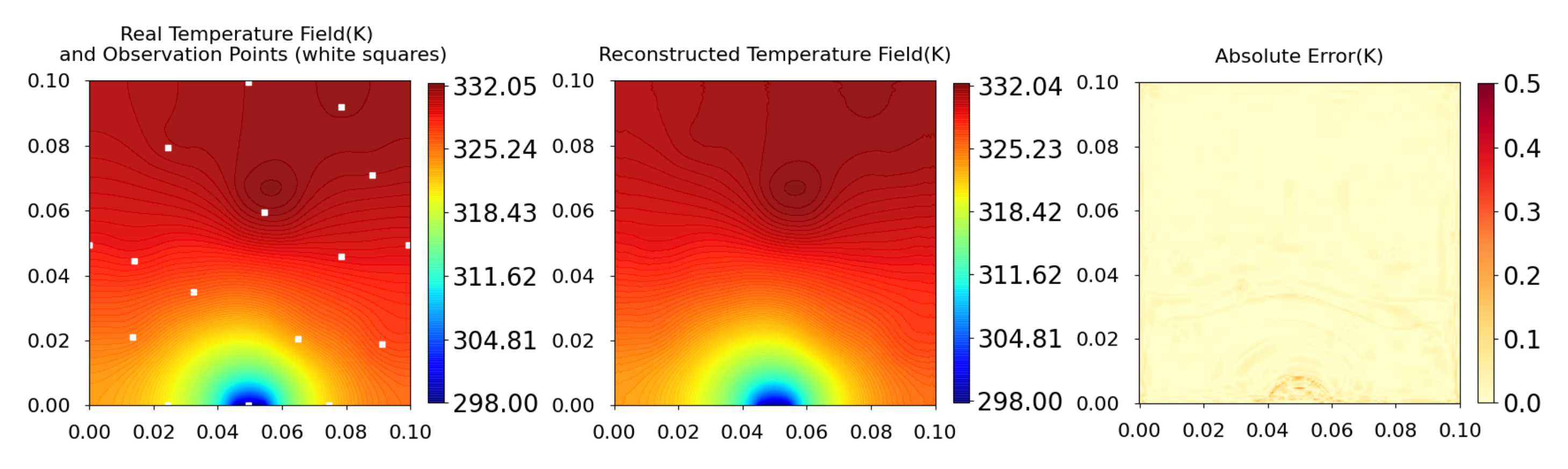}
		\label{fig:ps1_c1}
	}
	\caption{Performance with different observation points locations selection stratrgies for Case 1}
	\label{fig:locations_of_observation_points_c1}
\end{figure*}

\begin{figure*}[htbp]
	\centering
	\subfigure[Uniform selection, Case 2]{
		\includegraphics[width=1\linewidth]{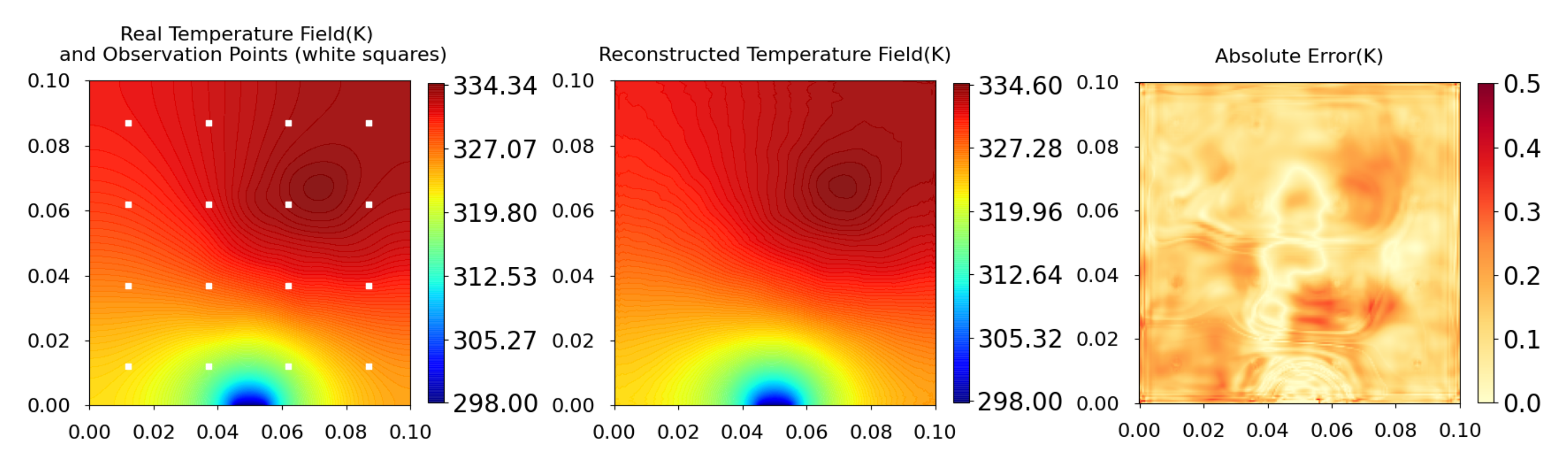}
		\label{fig:us_c2}
	}
	\subfigure[Random selection 1, Case 2]{
		\includegraphics[width=1\linewidth]{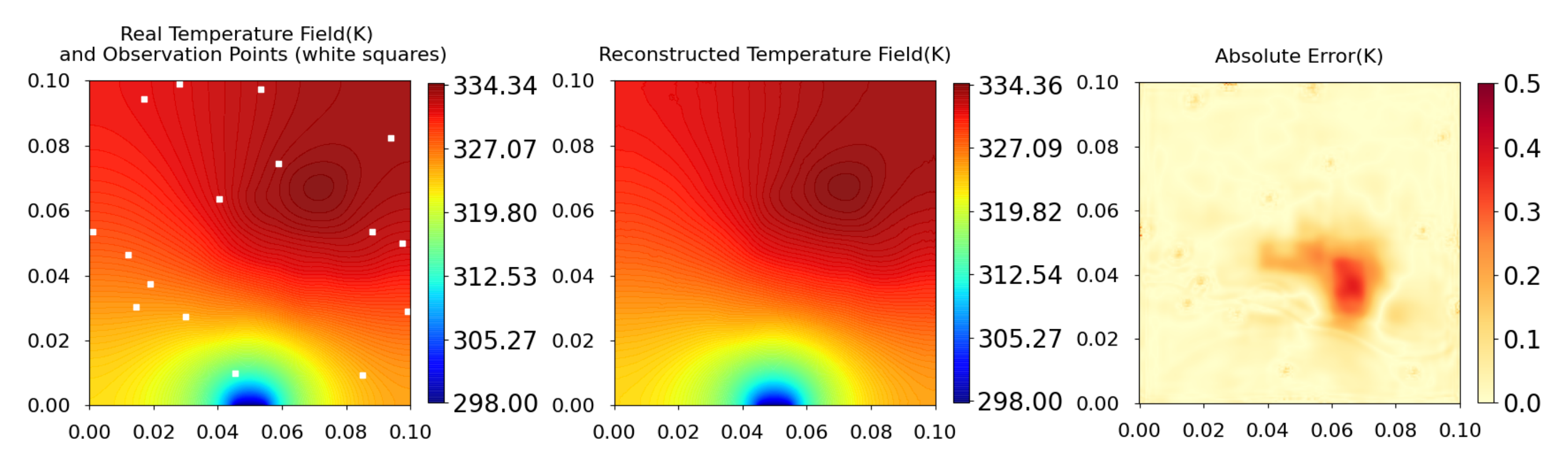}
		\label{fig:rs1_c2}
	}
	\subfigure[Random selection 2, Case 2]{
		\includegraphics[width=1\linewidth]{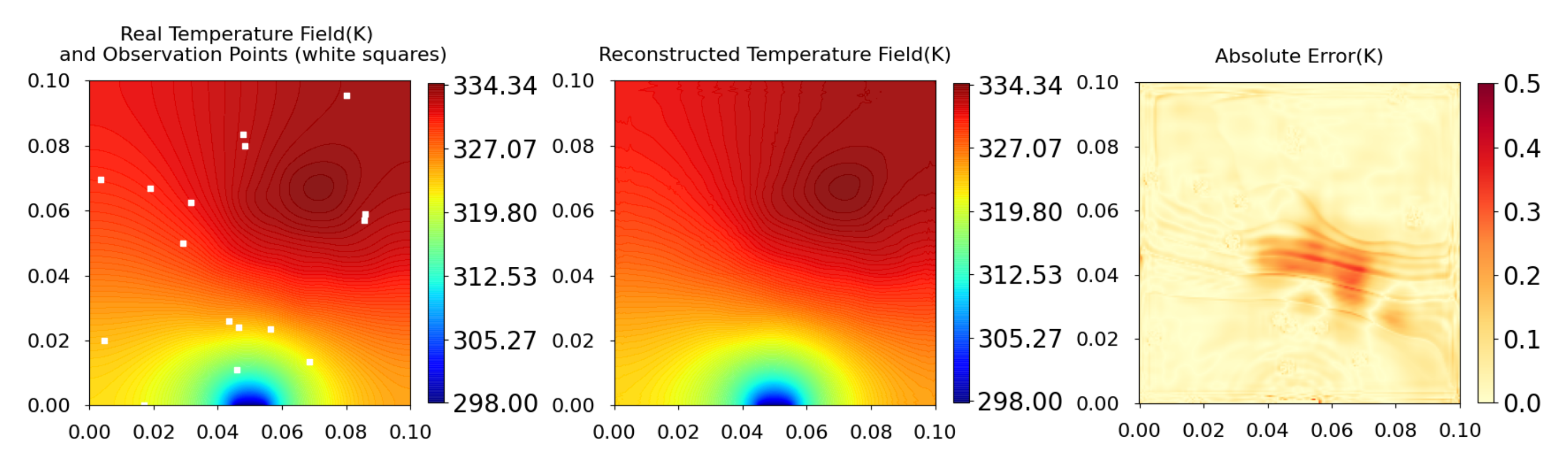}
		\label{fig:rs2_c2}
	}
	\subfigure[Physics-informed selection, Case 2]{
		\includegraphics[width=1\linewidth]{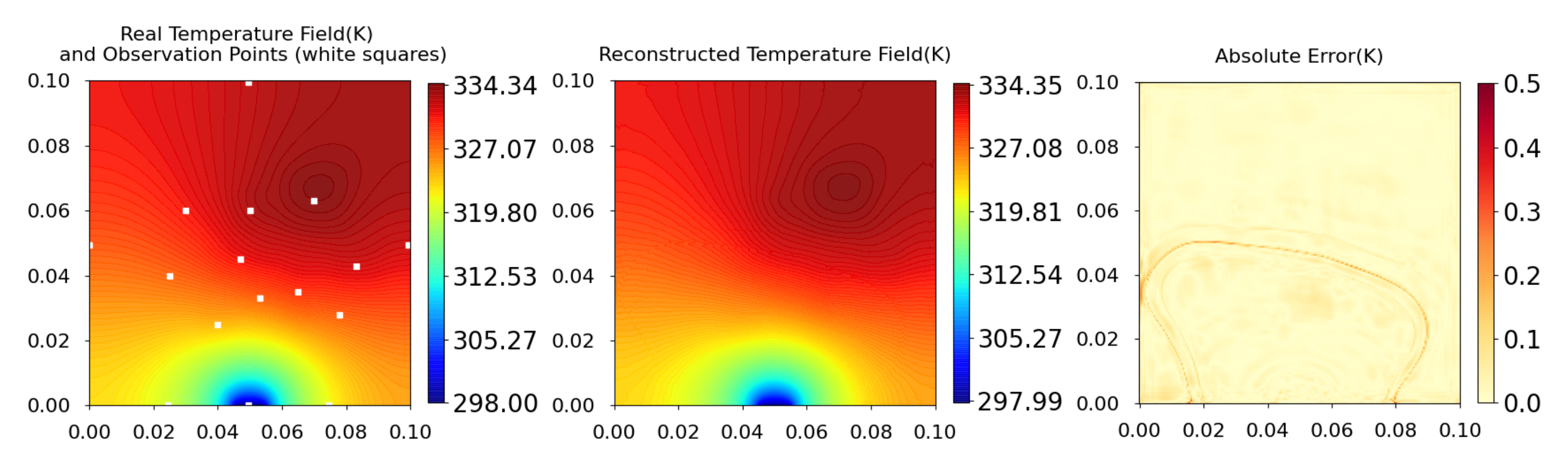}
		\label{fig:ps_c2}
	}
	\caption{Performance with different observation points locations selection stratrgies for Case 2}
	\label{fig:locations_of_observation_points_c2}
\end{figure*}

Table \ref{tab:3} lists the selection strategies and corresponding performances under different metrics. Fig. \ref{fig:locations_of_observation_points_c1} and Fig. \ref{fig:locations_of_observation_points_c2} shows some reconstruction examples with different locations of observation points. It can be found that the Uniform selection strategy and Physics-informed selection strategy perform better than Random selection strategy. Nevertheless, the reconstruction performances under two Random selection strategies are also good enough. This indicates that the proposed method can reconstruct the temperature field well for different locations of observation points. Furthermore, it is suggested that the locations of observation points are essential for the TFR. Setting the observation points on the components and boundaries is a simple and efficient approach. It should also be noted that the locations of the temperature observation points are restricted by the measurement technology and working conditions in engineering practice. To realize optimization of locations for observation points is an interesting and challenging task, which should be further investigated in the future.

\subsection{Comparisons with other methods}
\label{sec:5.3}

This work uses the global gaussian interpolation (\cite{TFRHSS}), kriging (\cite{kriging}), support vector regression (SVR) (\cite{Gaussian}) and neural network(NN) (\cite{Shallow}) as baselines. The number of test samples is set to 4000 for all methods and the number of training samples is set to 16000 for the supervised learning methods including the SVR, NN and the porposed method. 16 temperature observaton points is adopted for the TFR task. Table \ref{tab:4} shows the comparison results for different methods. For two cases, MAEs, CMAEs and MT-AEs of the proposed method are less than 0.1K, which is significantly lower than other methods by at least 10 times. This shows that the deep learning method generally outperforms other methods. On the other hand, MaxAEs of the former traditional methods exceed 10K for two cases, which are far more than that of the deep learning method. By contrast, the MaxAE of the proposed method is less than 0.5K due to the patchwise training approach. Since MaxAE evaluates the maximum absolute error with the ground-truth, it indicates that the deep learning method not only reconstruct the whole field precisely, but also get a higher precision estimation in the local regions than other methods. In a word, the proposed deep learning method has a better reconstructed performance and is more suitable for the TFR task.

\begin{table*}[htbp]
	\caption{The performances for different methods}
	\label{tab:4}
	\centering
	\begin{tabular}{ c c c c c c c }
		\hline\noalign{\smallskip}
		& Methods & NO & MAE & CMAE & MaxAE & MT-AE  \\
		\noalign{\smallskip}\hline\noalign{\smallskip}
		\multirow{6}{*}{Case 1}\centering & Kriging & 16 & 1.7261 & 0.3648 & 21.5361 & 2.4419 \\
		& Global Gaussian Interpolation & 16 & 1.0103 & 0.2117 & 18.9412 & 0.3482 \\
		& SVR & 16 & 1.7786 & 0.3416 & 26.3627 & 1.2651 \\
		& Neural Network & 16 & 1.5904 & 0.3298 & 15.7138 & 2.7387 \\
		& Proposed Method & 16 & \textbf{0.0113} & \textbf{0.0020} & \textbf{0.2602} & \textbf{0.0160}\\
		\hline\noalign{\smallskip}
		\multirow{6}{*}{Case 2}\centering & Kriging & 16 & 1.8547 & 0.3829 & 21.5361 & 2.8833 \\
		& Global Gaussian Interpolation & 16 & 0.9757 & 0.1767 & 18.6719 & 0.4044 \\
		& SVR & 16 & 1.8419 & 0.3618 & 26.4045 & 0.7165 \\
		& Neural Network & 16 & 0.7306 & 0.1613 & 14.7674 & 2.4431\\
		& Proposed Method & 16 & \textbf{0.0147} & \textbf{0.0030} & \textbf{0.4008} & \textbf{0.0358} \\
		\hline\noalign{\smallskip}
	\end{tabular}
\end{table*}

\section{Conclusions}
\label{sec:6}

In this paper, the TFR of electronic equipment is studied under the proposed deep learning method based on patchwise training. Treating the TFR as an image-to-image regression task, a patchwise training and inference framework consisting of an adaptive UNet and a shallow MLP is developed to establish the mapping from the observation to the temperature field. The adaptive UNet is utilized to reconstruct the whole temperature from limited observations. Meanwhile, the shallow MLP is designed to infer the temperature of the patch near the heatsink directly. In addition, gradient loss is introduced to improve the spatial smoothness of the reconstructed temperature field. Experiments demonstrate that the proposed method can reconstruct the temperature field accurately given sufficient training data. Especially the patchwise training approach improves the reconstruction accuracy through the shallow MLP. The maximum absolute errors of the reconstructed temperature field are less than 1K. Furthermore, by investigating cases under extreme power intensities, and different observation point numbers and locations, the generalization of the proposed method is validated. It needs only 16 temperature observation points to accurately reconstruct the temperature field and is not sensitive to the observation point locations. Besides, compared with the commonly used kriging method and other traditional surrogate model-based methods, the proposed method can provide better reconstruction performance with at least 10 times precision improvements.  

In future work, it is promising to study the TFR problem with more complex heat sources and boundary conditions. And a self-adaptive patchwise training approach will be investigated to solve the reconstruction of the region with large gradients better. Furthermore, it is worth investigating the time series problem about the reconstruction of the temperature field in real-time.

\section*{Conflict of interest statement}
On behalf of all authors, the corresponding author states that there is no conflict of interest.

\section*{Replication of results}
The code can be downloaded at: \url{https://github.com/pengxingwen/Reconstruction_pxw}.

\begin{acknowledgements}
The financial supports for this project are from the National Natural Science Foundation of China under Grant No.11725211 and 52005505. The authors would also like to thank Xianqi Chen at National University of Defense Technology for his help during the course of this work. 
\end{acknowledgements}

%
%

\bibliographystyle{spbasic}
\bibliography{references}

\begin{thebibliography}{44}
\providecommand{\natexlab}[1]{#1}
\providecommand{\url}[1]{{#1}}
\providecommand{\urlprefix}{URL }
\expandafter\ifx\csname urlstyle\endcsname\relax
  \providecommand{\doi}[1]{DOI~\discretionary{}{}{}#1}\else
  \providecommand{\doi}{DOI~\discretionary{}{}{}\begingroup
  \urlstyle{rm}\Url}\fi
\providecommand{\eprint}[2][]{\url{#2}}

\bibitem[{Aslan et~al.(2018)Aslan, Puskely, and Yarovoy}]{VP}
Aslan Y, Puskely J, Yarovoy AG (2018) Heat source layout optimization for
  two-dimensional heat conduction using iterative reweighted l1-norm convex
  minimization. International Journal of Heat and Mass Transfer 122:432–441

\bibitem[{Badrinarayanan et~al.(2017)Badrinarayanan, Kendall, and
  Cipolla}]{Segmentation}
Badrinarayanan V, Kendall A, Cipolla R (2017) Segnet: A deep convolutional
  encoder-decoder architecture for image segmentation. IEEE Transactions on
  Pattern Analysis and Machine Intelligence 39:2481--2495

\bibitem[{Chen et~al.(2020)Chen, Chen, Zhou, Zhang, and Yao}]{HSLO}
Chen X, Chen X, Zhou W, Zhang J, Yao W (2020) The heat source layout
  optimization using deep learning surrogate modeling. Structural and
  Multidisciplinary Optimization 62:1--22

\bibitem[{Chen et~al.(2021{\natexlab{a}})Chen, Gong, Zhao, Zhou, and
  Yao}]{TFRD}
Chen X, Gong Z, Zhao X, Zhou W, Yao W (2021{\natexlab{a}}) Tfrd: A benchmark
  dataset for research on temperature field reconstruction of heat-source
  systems. ArXiv abs/2108.08298

\bibitem[{Chen et~al.(2021{\natexlab{b}})Chen, Zhao, Gong, Zhang, Zhou, Chen,
  and Yao}]{Benchmark}
Chen X, Zhao X, Gong Z, Zhang J, Zhou W, Chen X, Yao W (2021{\natexlab{b}}) A
  deep neural network surrogate modeling benchmark for temperature field
  prediction of heat source layout. Science China Physics, Mechanics \&
  Astronomy 64(11):114611--1--30

\bibitem[{Dong et~al.(2014)Dong, Loy, He, and Tang}]{ImageSR}
Dong C, Loy CC, He K, Tang X (2014) Learning a deep convolutional network for
  image super-resolution. In: ECCV

\bibitem[{Erichson et~al.(2020)Erichson, Mathelin, Yao, Brunton, Mahoney, and
  Kutz}]{Shallow}
Erichson N, Mathelin L, Yao Z, Brunton SL, Mahoney MW, Kutz JN (2020) Shallow
  neural networks for fluid flow reconstruction with limited sensors.
  Proceedings of the Royal Society A 476

\bibitem[{Gong et~al.(2021{\natexlab{a}})Gong, Zhong, and Hu}]{Classification}
Gong Z, Zhong P, Hu W (2021{\natexlab{a}}) Statistical loss and analysis for
  deep learning in hyperspectral image classification. IEEE Transactions on
  Neural Networks and Learning Systems 32:322--333

\bibitem[{Gong et~al.(2021{\natexlab{b}})Gong, Zhou, Zhang, Peng, and
  Yao}]{TFRHSS}
Gong Z, Zhou W, Zhang J, Peng W, Yao W (2021{\natexlab{b}}) Physics-informed
  deep reversible regression model for temperature field reconstruction of
  heat-source systems. ArXiv abs/2106.11929

\bibitem[{Gu et~al.(2018)Gu, Wang, Kuen, Ma, Shahroudy, Shuai, Liu, Wang, Wang,
  Cai, and Chen}]{CNN}
Gu J, Wang Z, Kuen J, Ma L, Shahroudy A, Shuai B, Liu T, Wang X, Wang G, Cai J,
  Chen T (2018) Recent advances in convolutional neural networks. Pattern
  Recognit 77:354--377

\bibitem[{Guemes et~al.(2021)Guemes, Discetti, Ianiro, Sirmaçek, Azizpour, and
  Vinuesa}]{GAN}
Guemes A, Discetti S, Ianiro A, Sirmaçek B, Azizpour H, Vinuesa R (2021) From
  coarse wall measurements to turbulent velocity fields through deep learning.
  Physics of Fluids

\bibitem[{Holman(2002)}]{HeatTransfer}
Holman JP (2002) Heat Transfer. New York:McGraw-Hill

\bibitem[{Kingma and Ba(2015)}]{Adam}
Kingma DP, Ba J (2015) Adam: A method for stochastic optimization. CoRR
  abs/1412.6980

\bibitem[{Lei and Liu(2017)}]{ELM}
Lei J, Liu Q (2017) Three-dimensional temperature distribution reconstruction
  using the extreme learning machine. IET Signal Process 11:406--414

\bibitem[{Lei and Liu(2013)}]{GPOD}
Lei J, Liu S (2013) Temperature field reconstruction from the partial
  measurement data using the gappy proper orthogonal decomposition. Iet Science
  Measurement \& Technology 7:171--179

\bibitem[{Leon et~al.(2018)Leon, R{\'e}thor{\'e}, Dimitrov, Natarajan,
  S{\o}rensen, Graf, and Kim}]{PR}
Leon JPM, R{\'e}thor{\'e} PE, Dimitrov N, Natarajan A, S{\o}rensen JD, Graf PA,
  Kim T (2018) Uncertainty propagation through an aeroelastic wind turbine
  model using polynomial surrogates. Renewable Energy 119:910--922

\bibitem[{Li et~al.(2017)Li, Liu, Schlaberg, and Zhang}]{Iteration}
Li Y, Liu S, Schlaberg HI, Zhang J (2017) Temperature field reconstruction by
  acoustic based on newton-raphson regularization iteration. DEStech
  Transactions on Engineering and Technology Research

\bibitem[{Liu et~al.(2021)Liu, Li, Jing, Xie, and Zhang}]{Nano}
Liu T, Li Y, Jing Q, Xie Y, Zhang D (2021) Supervised learning method for the
  physical field reconstruction in a nanofluid heat transfer problem.
  International Journal of Heat and Mass Transfer 165:120684

\bibitem[{Miyauchi et~al.(2007)Miyauchi, Yu, Shibutani, and Shiratori}]{Stress}
Miyauchi H, Yu Q, Shibutani T, Shiratori M (2007) Evaluation technique for the
  failure life scatter of lead-free solder joints in electronic device. 2007
  13th International Workshop on Thermal Investigation of ICs and Systems
  (THERMINIC) pp 32--37

\bibitem[{Mohebbi et~al.(2017)Mohebbi, Sellier, and Rabczuk}]{Inverse}
Mohebbi F, Sellier M, Rabczuk T (2017) Inverse problem of simultaneously
  estimating the thermal conductivity and boundary shape. International Journal
  for Computational Methods in Engineering Science and Mechanics 18:166--181

\bibitem[{Morimoto et~al.(2021)Morimoto, Fukami, Zhang, and
  Fukagata}]{Generalization}
Morimoto M, Fukami K, Zhang K, Fukagata K (2021) Generalization techniques of
  neural networks for fluid flow estimation. Neural Computing and Applications

\bibitem[{Nakayama(1986)}]{Thermal}
Nakayama W (1986) Thermal management of electronic equipment: A review of
  technology and research topics. Applied Mechanics Reviews 39:1847--1868

\bibitem[{Narayana and Kumar(2016)}]{Transducer}
Narayana KVL, Kumar VN (2016) Development of an intelligent temperature
  transducer. IEEE Sensors Journal 16:4696--4703

\bibitem[{Popescu et~al.(2009)Popescu, Balas, Perescu-Popescu, and
  Mastorakis}]{MLP}
Popescu M, Balas VE, Perescu-Popescu L, Mastorakis NE (2009) Multilayer
  perceptron and neural networks. WSEAS Transactions on Circuits and Systems
  archive 8:579--588

\bibitem[{Protasov(2017)}]{Spline}
Protasov A (2017) Reconstruction of the thermal field image from measurements
  in separate points. 2017 IEEE Microwaves, Radar and Remote Sensing Symposium
  (MRRS) pp 89--92

\bibitem[{Raissi et~al.(2020)Raissi, Yazdani, and Karniadakis}]{PINN}
Raissi M, Yazdani A, Karniadakis GE (2020) Hidden fluid mechanics: Learning
  velocity and pressure fields from flow visualizations. Science 367:1026--1030

\bibitem[{Ronneberger et~al.(2015)Ronneberger, Fischer, and Brox}]{Unet}
Ronneberger O, Fischer P, Brox T (2015) U-net: Convolutional networks for
  biomedical image segmentation. In: International Conference on Medical Image
  Computing and Computer-assisted Intervention, pp 234--241

\bibitem[{Sch{\"a}fer and Finke(2008)}]{FEM2}
Sch{\"a}fer C, Finke E (2008) Shape optimisation by design of experiments and
  finite element methods—an application of steel wheels. Structural and
  Multidisciplinary Optimization 36:477--491

\bibitem[{Schulz et~al.(2018)Schulz, Speekenbrink, and Krause}]{Gaussian}
Schulz E, Speekenbrink M, Krause A (2018) A tutorial on gaussian process
  regression: Modelling, exploring, and exploiting functions. Journal of
  Mathematical Psychology

\bibitem[{Shelhamer et~al.(2017)Shelhamer, Long, and Darrell}]{FCN}
Shelhamer E, Long J, Darrell T (2017) Fully convolutional networks for semantic
  segmentation. IEEE Transactions on Pattern Analysis and Machine Intelligence
  39:640--651

\bibitem[{Srinivasan et~al.(2019)Srinivasan, Guastoni, Azizpour, Schlatter, and
  Vinuesa}]{RNN}
Srinivasan PAA, Guastoni L, Azizpour H, Schlatter P, Vinuesa R (2019)
  Predictions of turbulent shear flows using deep neural networks. Physical
  Review Fluids

\bibitem[{Sun and Wang(2020)}]{BNN}
Sun L, Wang JX (2020) Physics-constrained bayesian neural network for fluid
  flow reconstruction with sparse and noisy data. arXiv: Computational Physics

\bibitem[{Sun et~al.(2021)Sun, Zhang, Ji, and Qiu}]{IDW}
Sun Y, Zhang C, Ji H, Qiu J (2021) A temperature field reconstruction method
  for spacecraft leading edge structure with optimized sensor array. Journal of
  Intelligent Material Systems and Structures 32:2024--2038

\bibitem[{Tian et~al.(2019)Tian, Fang, and Wang}]{Kalman}
Tian N, Fang H, Wang Y (2019) 3-d temperature field reconstruction for a
  lithium-ion battery pack: A distributed kalman filtering approach. IEEE
  Transactions on Control Systems Technology 27:847--854

\bibitem[{Wu et~al.(2015)Wu, Kim, Han, Palczynska, and Gromala}]{Deformation}
Wu B, Kim DS, Han B, Palczynska A, Gromala PJ (2015) Thermal deformation
  analysis of automotive electronic control units subjected to passive and
  active thermal conditions. 2015 16th International Conference on Thermal,
  Mechanical and Multi-Physics Simulation and Experiments in Microelectronics
  and Microsystems pp 1--6

\bibitem[{Yan et~al.(2018)Yan, Shen, and Guo}]{SVR}
Yan C, Shen X, Guo F (2018) An improved support vector regression using least
  squares method. Structural and Multidisciplinary Optimization 57:2431--2445

\bibitem[{Yan et~al.(2011)Yan, Ma, Tan, and Ma}]{NN}
Yan R, Ma Y, Tan H, Ma J (2011) Temperature field reconstruction using
  artificial neural network. Proceedings of 2011 International Conference on
  Electronic \& Mechanical Engineering and Information Technology 4:2159--2163

\bibitem[{Yao et~al.(2012)Yao, Chen, Zhao, and van Tooren}]{RBF}
Yao W, Chen X, Zhao Y, van Tooren M (2012) Concurrent subspace width
  optimization method for rbf neural network modeling. IEEE Transactions on
  Neural Networks and Learning Systems 23:247--259

\bibitem[{Yeh(1995)}]{55}
Yeh LT (1995) Review of heat transfer technologies in electronic equipment.
  Journal of Electronic Packaging 117:333--339

\bibitem[{Zhang et~al.(2018)Zhang, Yao, Ye, and Chen}]{kriging}
Zhang Y, Yao W, Ye S, Chen X (2018) A regularization method for constructing
  trend function in kriging model. Structural and Multidisciplinary
  Optimization 59:1221--1239

\bibitem[{Zhao et~al.(2021)Zhao, Gong, Zhang, Yao, and Chen}]{Transfer}
Zhao X, Gong Z, Zhang J, Yao W, Chen X (2021) A surrogate model with data
  augmentation and deep transfer learning for temperature field prediction of
  heat source layout. Structural and Multidisciplinary Optimization

\bibitem[{Zhao et~al.(2019)Zhao, Zheng, Xu, and Wu}]{Detection}
Zhao ZQ, Zheng P, Xu St, Wu X (2019) Object detection with deep learning: A
  review. IEEE transactions on neural networks and learning systems
  30(11):3212--3232

\bibitem[{Zhu et~al.(2018)Zhu, Liu, Rosen, and Rosen}]{Reconstruction}
Zhu B, Liu JZ, Rosen BR, Rosen MS (2018) Image reconstruction by
  domain-transform manifold learning. Nature 555:487--492

\bibitem[{Zienkiewicz(1977)}]{FEM1}
Zienkiewicz O (1977) The finite element method

\end{thebibliography}

\end{document}